\documentclass[letterpaper]{article}
\usepackage{aaai2027}
\nocopyright
\usepackage[hyphens]{url}
\usepackage{graphicx}
\usepackage{natbib}
\usepackage{caption}
\usepackage{amsmath,amssymb,amsfonts}
\usepackage{booktabs}
\usepackage{multirow}
\title{STEAM: A Spatio-TEmporal Alignment Mixture-of-Experts Model with Hierarchical Pre-training for EEG Decoding}

\author{
  Zhu Chen\equalcontrib\textsuperscript{\rm 1,\rm 2},
  Dingkun Liu\equalcontrib\textsuperscript{\rm 1,\rm 2},
  Yuheng Chen\textsuperscript{\rm 1},
  Dongrui Wu\textsuperscript{\rm 1,\rm 2}\corresponding
}
\affiliations{
  \textsuperscript{\rm 1}Ministry of Education Key Laboratory of Image Processing and Intelligent Control, School of Artificial Intelligence and Automation, Huazhong University of Science and Technology, Wuhan 430074, China\\
  \textsuperscript{\rm 2}Zhongguancun Academy, Beijing 100094, China\\
  drwu09@gmail.com
}

\begin{document}

\maketitle

\begin{abstract}

Brain-computer interfaces (BCIs) have been widely used in motor rehabilitation, disease diagnosis, and other neural engineering scenarios. However, conventional neural signal decoding algorithms often suffer from limited generalizability and high adaptation costs, motivating recent interest in BCI foundation models. Existing approaches still struggle to jointly achieve general transferability, accurate decoding, and efficient downstream adaptation. We present STEAM, a hierarchical transfer framework that reconciles general-purpose representation learning with paradigm-specific specialization in EEG foundation models. The framework is instantiated as a dual-branch spatio-temporal encoder in which a shared soft mixture-of-experts (SSMoE) module aligns the spatial and temporal branches, allowing complementary representations to exchange information through a compact set of soft slots. Across seven downstream datasets and fourteen evaluation settings, STEAM attains the best average rank among the compared methods at a competitive inference cost measured in FLOPs. Building upon the Stage-I general initialization, the hierarchical pre-training strategy further specializes the model to a target paradigm without retraining from scratch, yielding consistent gains in paradigm-specific decoding accuracy.

\end{abstract}

\section{Introduction}

Brain--computer interfaces (BCIs) directly link neural activity to external devices~\cite{wolpaw2007brain,birbaumer2007brain}. Electroencephalography (EEG) is widely used for non-invasive recording because it is inexpensive, portable, and easy to deploy. Applications include motor rehabilitation, event-related potential detection, seizure monitoring, emotion recognition, and vigilance estimation~\cite{PhysioNet-chbmit-1.0.0,zheng2015investigating,zheng2017multimodal}. Yet EEG decoding remains difficult because of low signal-to-noise ratios, inter-subject variability, and heterogeneity in montages, sampling rates, preprocessing, and paradigms~\cite{lotte2007review}.

\begin{center}
\includegraphics[width=\columnwidth]{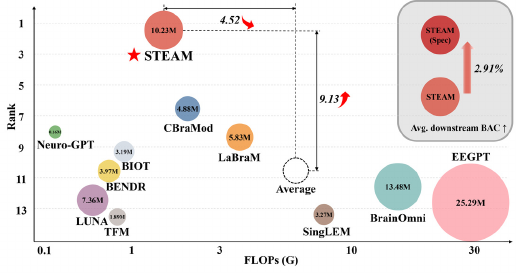}
\captionof{figure}{Comparison of EEG foundation models by overall rank and computational cost; bubble size indicates parameter count. For STEAM, only the Stage-I model is shown.}
\label{fig:fm_cost_rank}
\end{center}

Deep neural networks (DNNs), particularly convolutional architectures, have been widely adopted for end-to-end EEG decoding by learning temporal, spectral, and spatial patterns~\cite{schirrmeister2017deep,lawhern2018eegnet,craik2019deep}.
However, conventional models are often trained from scratch for individual datasets or acquisition settings, limiting their transferability across subjects, electrode configurations, recording devices, paradigms, and tasks~\cite{Liu2025sdda}.

Large-scale self-supervised pre-training has driven the development of EEG foundation models (EEG-FMs)~\cite{kostas2021bendr,jiang2024labram,wang2024eegpt}. General-purpose EEG-FMs target broad transfer across paradigms and tasks, whereas paradigm-specific models target cross-subject and cross-dataset transfer within a specific paradigm~\cite{liu2026mirepnet}. The former favor breadth, while the latter leverage paradigm-specific neurophysiology for stronger discrimination. Nevertheless, existing EEG-FMs still face three coupled challenges.

First, cross-paradigm pre-training does not consistently translate into strong downstream decoding performance. Benchmarks show that general-purpose EEG-FMs can underperform specialist models on several tasks~\cite{liu2026bench}. This gap calls for transferable EEG representations that explicitly preserve channel-wise spatial dependencies, temporal dynamics, and frequency-dependent patterns, rather than prematurely collapsing these heterogeneous cues into a single stream.

Second, broad transferability and paradigm-specific discrimination are difficult to reconcile. Cross-paradigm pre-training provides a reusable initialization but may weaken target-specific cues, whereas training a separate foundation model for each paradigm is costly and impractical, particularly when data are limited. EEG-FMs should therefore support efficient specialization while preserving general pre-trained knowledge.

Third, downstream adaptation remains data- and parameter-intensive. EEG-FMs often require full-parameter fine-tuning or target-domain post-training and may perform poorly when only a few labeled trials are available. Such protocols limit the efficient reuse of a pre-trained backbone across subjects, electrode configurations, and tasks~\cite{houlsby2019adapter,lester2021power}. Accordingly, the central problem is to jointly achieve robust cross-paradigm transfer, effective paradigm-level specialization, and parameter-efficient downstream adaptation.

To address these issues, we propose STEAM, a Spatio-TEmporal Alignment Mixture-of-Experts framework for transferable EEG decoding. The main contributions are summarized as follows:

\begin{enumerate}

\item
We propose a general-to-paradigm hierarchical pre-training framework for EEG foundation models that progressively specializes general EEG representations for paradigm-specific decoding. The framework first learns a reusable backbone through masked reconstruction and spatio-temporal contrastive alignment, and then continues pre-training on multiple datasets within a target paradigm using supervised objectives.

\item
We introduce STEAM, a dual-branch spatio-temporal architecture tailored for hierarchical EEG representation learning. STEAM preserves complementary channel-centric and temporal-segment-centric representations, while a shared soft mixture-of-experts module enables high-level cross-branch interaction through compact soft slots. Frequency-aware gated attention further incorporates spectral priors into both representation streams.

\item
We present a lightweight downstream adaptation strategy that updates only the spatial and temporal embedding modules and the task head, accounting for approximately 5.1\% of the model parameters. Extensive experiments across seven downstream datasets and fourteen evaluation settings show that STEAM achieves the best average rank at a relatively low inference cost (see Fig.~\ref{fig:fm_cost_rank}). Paradigm-specific specialization further improves target-paradigm decoding, while lightweight adaptation remains competitive with full fine-tuning on most classification tasks.

\end{enumerate}

\section{Related Work}

\subsection{Task-Specific EEG Decoding Models}

Before end-to-end networks, EEG decoding used handcrafted temporal, spectral, and spatial features. For motor imagery, common spatial patterns (CSP) and variants extracted variance-based spatial features distinguishing sensorimotor states~\cite{ramoser2000optimal,blankertz2008optimizing}. Although interpretable, their performance depends on careful preprocessing and feature extraction.

Convolutional networks enabled end-to-end EEG representation learning. ShallowConvNet and DeepConvNet~\cite{schirrmeister2017deep} use temporal and spatial convolutions to capture rhythmic-power patterns and increasingly abstract spatiotemporal features, respectively. EEGNet~\cite{lawhern2018eegnet} uses depthwise and separable convolutions for a compact architecture across tasks.

Recent architectures combine convolutional biases and self-attention. EEG-Conformer~\cite{song2022eeg} couples local convolutions with global Transformer modeling; EEG-Deformer~\cite{ding2024eeg} and DBConformer~\cite{wang2025dbconformer} use denser or dual-branch mechanisms for spatiotemporal interaction. Despite strong in-domain performance, these task-specific models are trained from scratch for fixed tasks and acquisition settings, limiting reuse across subjects, devices, datasets, and paradigms.

\subsection{EEG Foundation Models}

EEG foundation models (EEG-FMs) learn reusable representations from large-scale recordings for transfer across subjects, datasets, acquisition systems, and tasks. BENDR~\cite{kostas2021bendr} uses contrastive learning over contextualized EEG representations, while BIOT~\cite{yang2023biot} unifies heterogeneous biosignals for cross-dataset and cross-signal learning. LaBraM~\cite{jiang2024labram} combines channel-wise patching, neural-spectrum quantization, and masked token prediction, and EEGPT~\cite{wang2024eegpt} combines masked modeling with spatio-temporal alignment. Structurally informed models include CBraMod~\cite{wang2025cbramod}, with criss-cross spatial and temporal dependencies; CSBrain~\cite{zhou2026csbrain}, with cross-scale representations; EEGMamba~\cite{wang2025eegmamba}, with selective state-space modeling of long sequences; and NeuroLM~\cite{jiang2025neurolm}, with neural tokenization and language models for multitask EEG inference.

General-purpose EEG-FMs target cross-paradigm transfer; paradigm-specific models target cross-subject and cross-dataset transfer within a task family. MIRepNet~\cite{liu2026mirepnet} combines neurophysiological priors, channel alignment, masked reconstruction, and supervised classification for few-shot motor-imagery adaptation. SleepFM~\cite{thapa2024sleepfm} learns sleep representations from multimodal polysomnography. These models show that paradigm priors and consistent labels strengthen transfer within a physiological domain.

This trade-off motivates STEAM's general-to-paradigm hierarchical pre-training, which specializes a reusable backbone without training separate models from scratch.

\section{Method}

\begin{figure*}[t]
    \centering
    \includegraphics[width=0.98\textwidth]{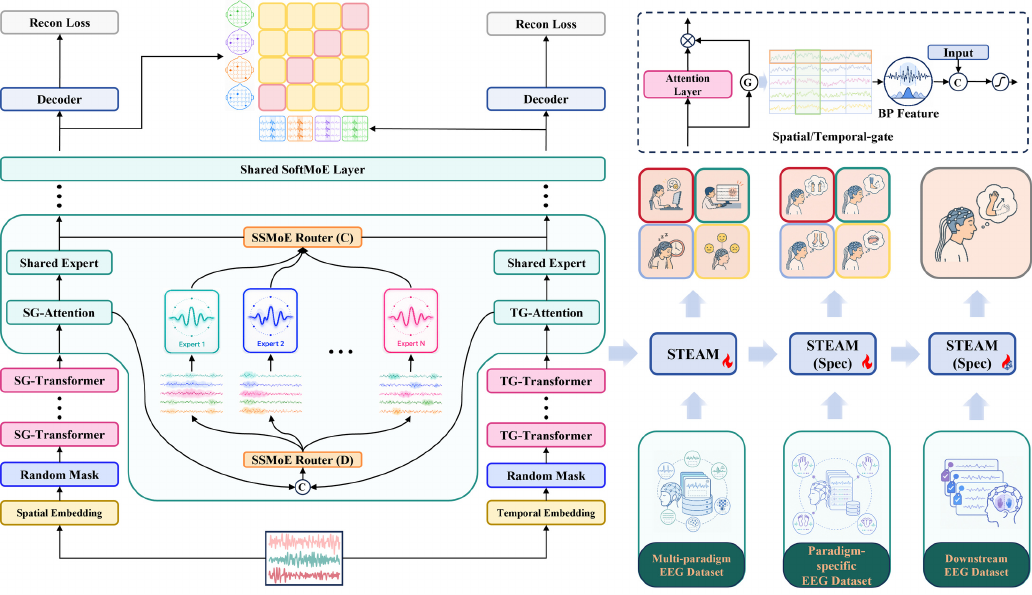}
    \caption{Overview of the proposed STEAM framework. \textbf{Left:} The overall model architecture, in which EEG signals are processed by spatial and temporal embedding branches, randomly masked for self-supervised reconstruction, and encoded by branch-specific Transformer blocks. The two branches are further coupled through the shared SSMoE module, where soft routing aggregates spatial and temporal tokens into expert slots and redistributes the fused representations back to the corresponding branches. This architecture supports a contrastive pre-training framework that aligns sample-level spatial and temporal representations while preserving local reconstruction targets. \textbf{Upper right:} The frequency-aware gated attention module, which extracts band-power descriptors from EEG segments and uses them to modulate layerwise attention responses. \textbf{Lower right:} The training and application pipeline, including multi-paradigm EEG pre-training, paradigm-specific continual pre-training, and downstream task adaptation with lightweight fine-tuning.}
    \label{fig:steam_overview}
\end{figure*}

To reconcile general EEG representation learning with paradigm-specific specialization, we propose a general-to-paradigm hierarchical pre-training framework that progressively adapts transferable representations to a target paradigm without retraining from scratch.
Building on this framework, STEAM introduces a dual-branch spatio-temporal encoder with shared soft mixture-of-experts interaction and frequency-aware gated attention.

\subsection{Architecture of STEAM}

\subsubsection{Dual-Branch Architecture}

Discriminative EEG patterns reflect spatial distributions across channels and temporal dynamics within trials. Accordingly, STEAM uses a dual-branch encoder to construct complementary channel-centric and temporal-segment-centric representations. The spatial embedding module $E_s$ produces channel-wise tokens:
\begin{equation}
    \mathbf{S}^{(0)}=E_s(\mathbf{X})
    \in\mathbb{R}^{B\times C\times d},
\end{equation}
where $d$ is the hidden dimension. Each spatial token corresponds to an input channel, enabling attention layers to model cross-channel dependencies. In parallel, the temporal embedding module $E_t$ produces temporal-segment tokens:
\begin{equation}
    \mathbf{T}^{(0)}=E_t(\mathbf{X})
    \in\mathbb{R}^{B\times P\times d},
\end{equation}
where $P$ denotes the number of temporal segments. At layer $l$, the branch-specific blocks are updated independently to obtain the pre-fusion representations $\bar{\mathbf{S}}^{(l)}$ and $\bar{\mathbf{T}}^{(l)}$ before cross-branch interaction. Delayed interaction preserves cross-channel modeling in the spatial branch and local-to-global dynamics in the temporal branch.

\subsubsection{Spatio-Temporal SSMoE Architecture}

To couple the otherwise independent branches, we insert a shared soft mixture-of-experts (SSMoE) module at selected intermediate and upper layers, following soft MoE routing principles~\cite{puigcerver2024sparse}. Let $\bar{\mathbf{S}}^{(l)}$ and $\bar{\mathbf{T}}^{(l)}$ denote the branch-specific representations after self-attention and feed-forward updates at layer $l$. SSMoE routes tokens from both branches through a shared expert pool, enabling information exchange while retaining branch-specific pathways.

The two token sequences are first concatenated as
\begin{equation}
    \mathbf{Z}^{(l)}=[\bar{\mathbf{S}}^{(l)};\bar{\mathbf{T}}^{(l)}]
    \in\mathbb{R}^{B\times N\times d},\quad N=C+P.
\end{equation}
Let $M$ denote the total number of expert slots. A learnable routing matrix $\boldsymbol{\Phi}^{(l)}\in\mathbb{R}^{d\times M}$ produces the
token-to-slot affinity scores:
\begin{equation}
    \mathbf{R}^{(l)}=\mathbf{Z}^{(l)}\boldsymbol{\Phi}^{(l)}
    \in\mathbb{R}^{B\times N\times M}.
\end{equation}
Here, $R^{(l)}_{b,n,m}$ measures the affinity between the $n$-th token and the $m$-th slot. The dispatch and combine weights are computed as
\begin{equation}
    \mathbf{D}^{(l)}=\operatorname{softmax}_{N}(\mathbf{R}^{(l)}),\quad
    \mathbf{C}^{(l)}=\operatorname{softmax}_{M}(\mathbf{R}^{(l)}).
\end{equation}
The dispatch weights are normalized over the $N$ input tokens per slot, whereas the combine weights are normalized over the $M$ slots per token. The slot representations and expert-transformed outputs are
\begin{equation}
\begin{aligned}
    \mathbf{U}^{(l)}_{b,m,:}
    &=\sum_{n=1}^{N}D^{(l)}_{b,n,m}\mathbf{Z}^{(l)}_{b,n,:},\\
    \mathbf{V}^{(l)}_{b,m,:}
    &=f^{(l)}_{e(m)}\bigl(\mathbf{U}^{(l)}_{b,m,:}\bigr),
\end{aligned}
\end{equation}
where $e(m)$ denotes the expert associated with slot $m$. Each original token then receives a token-specific mixture of the transformed slots:
\begin{equation}
    \mathbf{Y}^{(l)}_{b,n,:}
    =\sum_{m=1}^{M}C^{(l)}_{b,n,m}\mathbf{V}^{(l)}_{b,m,:}.
\end{equation}
Finally, the output sequence is divided according to the original branch boundaries and added to the corresponding pre-fusion representations:
\begin{equation}
\begin{aligned}
    [\mathbf{Y}^{(l)}_s;\mathbf{Y}^{(l)}_t]&=\mathbf{Y}^{(l)},\\
    \mathbf{S}^{(l)}&=\bar{\mathbf{S}}^{(l)}+\mathbf{Y}^{(l)}_s,\quad
    \mathbf{T}^{(l)}=\bar{\mathbf{T}}^{(l)}+\mathbf{Y}^{(l)}_t.
\end{aligned}
\end{equation}

Each slot aggregates global information from both branches, while token-specific combine weights redistribute the transformed slots without disrupting token organization. This compact bottleneck captures spatial, temporal, and joint factors; unlike top-$k$ routing, soft routing lets each token interact with multiple slots, which suits distributed EEG patterns under low signal-to-noise ratios.

\subsubsection{Frequency-Aware Gated Attention}

EEG recordings typically exhibit low signal-to-noise ratios, making purely data-driven attention sensitive to noise-related components. To introduce an explicit frequency-aware prior, STEAM modulates the attention output in each branch encoder layer using band-power descriptors~\cite{pfurtscheller1999event}. Given a token-associated signal segment $u$, its log-power feature in the $k$-th frequency band $\Omega_k$ is defined as
\begin{equation}
    b_k(u)=\log\left(1+\frac{1}{|\Omega_k|}
    \sum_{f\in\Omega_k}|\operatorname{FFT}(u)_f|^2\right).
\end{equation}
Applying this operation to the signal support associated with each spatial and temporal token yields $\mathbf{F}_s\in\mathbb{R}^{B\times C\times K}$ and $\mathbf{F}_t\in\mathbb{R}^{B\times P\times K}$, respectively, where $K$ is the number of frequency bands.

For branch $q\in\{s,t\}$, let $\mathbf{A}^{(l)}_q\in\mathbb{R}^{B\times N_q\times d}$ denote the attention output at layer $l$, where $N_s=C$ and $N_t=P$. The frequency-guided scalar gate is computed as
\begin{equation}
    \begin{aligned}
    \mathbf{G}^{(l)}_q
    &=\sigma\Bigl(
    \phi\Bigl(
    [\operatorname{LN}(\mathbf{A}^{(l)}_q);
    \operatorname{LN}(\mathbf{F}_q)]\\
    &\qquad\mathbf{W}^{(l)}_{q,1}
    +\mathbf{b}^{(l)}_{q,1}\Bigr)
    \mathbf{W}^{(l)}_{q,2}+\mathbf{b}^{(l)}_{q,2}\Bigr),
    \end{aligned}
\end{equation}
where $[\cdot;\cdot]$ denotes feature-wise concatenation, $\mathbf{W}^{(l)}_{q,1}\in\mathbb{R}^{(d+K)\times d}$ and $\mathbf{W}^{(l)}_{q,2}\in\mathbb{R}^{d\times1}$ are learnable projections, $\phi$ denotes GELU, and $\sigma$ denotes sigmoid. Thus, $\mathbf{G}^{(l)}_q\in\mathbb{R}^{B\times N_q\times1}$ assigns each token a scalar weight, which reweights the attention output through broadcasting before the residual connection:
\begin{equation}
    \widetilde{\mathbf{A}}^{(l)}_q=
    \mathbf{G}^{(l)}_q\odot\mathbf{A}^{(l)}_q.
\end{equation}

Unlike late fusion at the prediction head, this mechanism injects frequency-domain evidence throughout the encoder, preserving self-attention's contextual modeling while introducing a spectral inductive bias into both branches.

\subsection{Hierarchical pre-training and Lightweight Adaptation}

\subsubsection{Masked Reconstruction and Cross-View Contrastive Learning}

The spatial and temporal branches should preserve view-specific local structures while sharing trial-level semantics. STEAM therefore combines branch-specific masked reconstruction with cross-view contrastive learning. For the $i$-th EEG trial, the final spatial and temporal token sequences are pooled as
\begin{equation}
    \mathbf{p}_{s,i}=\operatorname{Pool}(\mathbf{S}^{(L)}_i),\quad
    \mathbf{p}_{t,i}=\operatorname{Pool}(\mathbf{T}^{(L)}_i).
\end{equation}
Two branch-specific projection heads $\pi_s$ and $\pi_t$ then produce
$\ell_2$-normalized contrastive representations:
\begin{equation}
    \mathbf{z}_{q,i}=
    \frac{\pi_q(\mathbf{p}_{q,i})}
    {\|\pi_q(\mathbf{p}_{q,i})\|_2},\quad q\in\{s,t\}.
\end{equation}
The spatial and temporal representations of the same trial form a positive pair, whereas representations from different trials in the mini-batch are treated as negative pairs. The symmetric InfoNCE loss is
\begin{equation}
\begin{aligned}
    \mathcal{L}_{\mathrm{con}}
    =-\frac{1}{2B}\sum_{i=1}^{B}\Bigg[
    &\log
    \frac{\exp(\mathbf{z}_{s,i}^{\top}\mathbf{z}_{t,i}/\tau)}
    {\sum_{j=1}^{B}\exp(\mathbf{z}_{s,i}^{\top}\mathbf{z}_{t,j}/\tau)}
    \\
    +&\log
    \frac{\exp(\mathbf{z}_{t,i}^{\top}\mathbf{z}_{s,i}/\tau)}
    {\sum_{j=1}^{B}\exp(\mathbf{z}_{t,i}^{\top}\mathbf{z}_{s,j}/\tau)}
    \Bigg],
\end{aligned}
\end{equation}
where $\tau>0$ is the temperature parameter.

In parallel, lightweight branch-specific decoders reconstruct the randomly masked token targets. Let $\mathcal{M}_q$ denote the masked token indices of branch $q$, and let $\mathbf{v}_{q,r}$ and $\hat{\mathbf{v}}_{q,r}$ denote the target and reconstructed tokens, respectively. The reconstruction loss is
\begin{equation}
    \mathcal{L}_{\mathrm{rec}}=
    \sum_{q\in\{s,t\}}\frac{1}{|\mathcal{M}_q|}
    \sum_{r\in\mathcal{M}_q}
    \|\hat{\mathbf{v}}_{q,r}-\mathbf{v}_{q,r}\|_2^2.
\end{equation}
To prevent reconstruction collapse, a diversity loss $\mathcal{L}_{\mathrm{div}}$ discourages degenerate reconstructions. Masked reconstruction preserves view-specific local information, while contrastive learning aligns trial-level branch representations. Joint optimization retains complementary spatial and temporal structures with consistent global semantics. The reconstruction decoders and contrastive heads are discarded after pre-training.

\subsubsection{General-to-Paradigm Hierarchical Pre-training}

Broad EEG pre-training provides a reusable initialization, but a mixed-data objective may not preserve target-paradigm discriminative cues. Conversely, training a separate foundation model for every paradigm incurs substantial costs. STEAM therefore adopts a two-stage general-to-paradigm pre-training strategy.

In Stage-I, the model is pre-trained on large-scale unlabeled EEG data using masked reconstruction and cross-view contrastive alignment:
\begin{equation}
    \mathcal{L}_{\mathrm{I}}=\mathcal{L}_{\mathrm{rec}}
    +\alpha\mathcal{L}_{\mathrm{div}}
    +\lambda\mathcal{L}_{\mathrm{con}}.
\end{equation}
This stage establishes a reusable spatio-temporal initialization. In Stage-II, training continues on multiple datasets from a target paradigm, with an additional supervised objective:
\begin{equation}
    \mathcal{L}_{\mathrm{II}}=\mathcal{L}_{\mathrm{rec}}
    +\alpha\mathcal{L}_{\mathrm{div}}
    +\lambda\mathcal{L}_{\mathrm{con}}
    +\mu\mathcal{L}_{\mathrm{cls}},
\end{equation}
where $\alpha$, $\lambda$, and $\mu$ control the contributions of the diversity, contrastive, and supervised classification objectives, respectively. When the Stage-II datasets have different label spaces, $\mathcal{L}_{\mathrm{cls}}$ is computed using dataset-specific classification heads and aggregated across datasets. Stage-II therefore refines the general initialization toward target-paradigm decision boundaries without training a separate backbone from scratch.

\subsubsection{Lightweight Downstream Adaptation}

For downstream transfer, we consider both full fine-tuning and a lightweight adaptation strategy motivated by parameter-efficient transfer learning. Among the pre-trained parameters, only the spatial and temporal embedding modules are updated, while the remaining encoder parameters are frozen. A task-specific prediction head $g_{\psi}$ is initialized for each downstream dataset.

Let $\theta_{\mathrm{emb}}=\{\theta_s,\theta_t\}$ denote the trainable parameters of the two embedding modules, and let $F_{\theta_0}$ denote the frozen spatio-temporal encoder. For the $i$-th downstream trial, its fused representation is
\begin{equation}
    \mathbf{h}_i=
    \operatorname{FusePool}\left(
    F_{\theta_0}(E_{\theta_{\mathrm{emb}}}(\mathbf{x}_i))\right),
\end{equation}
where $E_{\theta_{\mathrm{emb}}}$ produces the spatial and temporal input tokens, and $\operatorname{FusePool}$ pools and combines the final outputs of the two branches. The downstream objective is
\begin{equation}
    \min_{\theta_{\mathrm{emb}},\psi}
    \frac{1}{n}\sum_{i=1}^{n}
    \ell_{\mathrm{down}}
    \left(g_{\psi}(\mathbf{h}_i),y_i\right),
\end{equation}
where $n$ is the number of labeled downstream trials, and $\ell_{\mathrm{down}}$ denotes the classification or regression loss. Restricting optimization to the embedding modules and task head reduces the number of updated parameters and the associated optimizer-state storage.

\section{Experiment}
\subsection{Datasets}

STEAM is pre-trained in two stages on disjoint general-purpose and paradigm-specific EEG corpora. Stage-I uses the Temple University Hospital EEG Corpus (TUH)~\cite{obeid2016temple} to learn a general spatio-temporal initialization. Stage-II performs paradigm-specific continual pre-training on seven MI datasets, including BNCI2014002~\cite{steyrl2016random}, Cho2017~\cite{cho2017eeg}, Dreyer2023~\cite{dreyer2023large}, Lee2019~\cite{lee2019eeg},  Weibo2014~\cite{yi2014evaluation}, Zhou2016~\cite{zhou2016fully}, and PhysioNetMI~\cite{schalk2004bci2000}, or on SEED-V~\cite{liu2021comparing} and DEAP~\cite{koelstra2011deap} for emotion recognition, yielding the corresponding STEAM-Spec variants. Further details are provided in Appendix~B.

The evaluation includes seven datasets spanning six applications: BNCI2014001~\cite{tangermann2012review} and BNCI2015001~\cite{6177271} for MI, BNCI2014009~\cite{arico2014influence} for P300 detection, CHB-MIT~\cite{PhysioNet-chbmit-1.0.0} for seizure detection, EEGMAT~\cite{zyma2019electroencephalograms} for cognitive-state recognition, SEED~\cite{zheng2015investigating} for emotion recognition, and SEED-VIG~\cite{zheng2017multimodal} for vigilance estimation. Their variations in subjects, channel configurations, recording durations, label spaces, and prediction types form a heterogeneous transfer benchmark.

\begin{table*}[!tb]
\centering
\begingroup
\small
\setlength{\tabcolsep}{2.0pt}
\begin{tabular}{c|c|ccccccc}
\toprule
Model Type & Approach & BNCI2014001 & BNCI2015001 & BNCI2014009 & CHB-MIT & SEED & EEGMAT & SEED-VIG \\
\midrule
\multicolumn{9}{l}{\textit{Cross-subject (LOSO)}} \\
\multirow{6}{*}{\begin{tabular}{@{}c@{}}Specialist\\models\end{tabular}} & EEGNet & $44.97_{\pm0.57}$ & $63.40_{\pm1.32}$ & $78.39_{\pm0.44}$ & $77.36_{\pm0.54}$ & $48.57_{\pm1.34}$ & $66.60_{\pm0.63}$ & $0.2561_{\pm0.0092}$ \\
 & ShallowConv & $44.80_{\pm0.50}$ & $63.89_{\pm0.80}$ & $78.05_{\pm0.62}$ & $\underline{80.18}_{\pm0.22}$ & $53.41_{\pm0.12}$ & $72.22_{\pm1.24}$ & $0.2290_{\pm0.0029}$ \\
 & LMDA & $46.80_{\pm0.31}$ & $61.40_{\pm0.94}$ & $\underline{78.45}_{\pm0.46}$ & $77.47_{\pm0.41}$ & $50.12_{\pm0.43}$ & $67.47_{\pm1.21}$ & $0.2389_{\pm0.0029}$ \\
 & CNN-T & $39.15_{\pm0.56}$ & $59.64_{\pm1.41}$ & $61.50_{\pm1.71}$ & $76.34_{\pm0.61}$ & $44.56_{\pm1.40}$ & $70.77_{\pm2.49}$ & $0.2556_{\pm0.0140}$ \\
 & Deformer & $41.53_{\pm0.67}$ & $63.06_{\pm1.17}$ & $76.89_{\pm0.14}$ & $79.77_{\pm0.14}$ & $51.05_{\pm0.97}$ & $71.73_{\pm0.37}$ & $0.2512_{\pm0.0053}$ \\
 & Conformer & $41.64_{\pm1.23}$ & $59.10_{\pm2.14}$ & $62.22_{\pm0.97}$ & $78.58_{\pm0.56}$ & $48.76_{\pm1.23}$ & $70.49_{\pm0.93}$ & $0.2405_{\pm0.0044}$ \\
\cmidrule(lr){1-9}
\multirow{13}{*}{\begin{tabular}{@{}c@{}}Foundation\\models\end{tabular}} & BENDR & $51.11_{\pm0.25}$ & $62.68_{\pm0.49}$ & $73.46_{\pm0.28}$ & $75.50_{\pm0.93}$ & $52.50_{\pm0.85}$ & $54.32_{\pm0.71}$ & $0.2412_{\pm0.0025}$ \\
 & BIOT & $34.27_{\pm0.93}$ & $63.94_{\pm1.20}$ & $58.14_{\pm0.33}$ & $74.85_{\pm0.33}$ & $49.04_{\pm0.90}$ & $70.77_{\pm1.49}$ & $0.2374_{\pm0.0033}$ \\
 & LaBraM & $46.93_{\pm1.43}$ & $64.14_{\pm1.03}$ & $70.31_{\pm0.24}$ & $70.87_{\pm0.59}$ & $52.23_{\pm0.92}$ & $65.74_{\pm1.61}$ & $\mathbf{0.2281}_{\pm0.0035}$ \\
 & Neuro-GPT & $46.97_{\pm0.71}$ & $60.62_{\pm1.63}$ & $75.97_{\pm0.53}$ & $73.27_{\pm0.27}$ & $49.67_{\pm0.38}$ & $\underline{72.62}_{\pm1.35}$ & $0.2509_{\pm0.0055}$ \\
 & EEGPT & $32.24_{\pm1.45}$ & $59.88_{\pm1.39}$ & $62.77_{\pm1.85}$ & $66.91_{\pm2.89}$ & $48.74_{\pm3.41}$ & $58.02_{\pm1.02}$ & $0.2402_{\pm0.0025}$ \\
 & CBraMod & $53.03_{\pm0.22}$ & $63.47_{\pm0.36}$ & $77.30_{\pm0.28}$ & $74.23_{\pm0.19}$ & ${53.61}_{\pm0.61}$ & $68.43_{\pm0.72}$ & $0.2718_{\pm0.0019}$ \\
 & TFM & $32.02_{\pm0.66}$ & $55.35_{\pm1.46}$ & $53.10_{\pm0.48}$ & $63.46_{\pm0.60}$ & $36.66_{\pm0.26}$ & $63.02_{\pm1.95}$ & $\underline{0.2283}_{\pm0.0026}$ \\
 & BrainOmni & $41.58_{\pm0.80}$ & $61.88_{\pm0.30}$ & $70.48_{\pm0.23}$ & -- & $38.02_{\pm0.03}$ & $57.50_{\pm0.80}$ & $0.2434_{\pm0.0003}$ \\
 & SingLEM & $30.57_{\pm0.10}$ & $54.47_{\pm0.62}$ & $71.98_{\pm0.56}$ & $60.78_{\pm1.82}$ & $50.16_{\pm1.48}$ & $49.85_{\pm0.16}$ & $0.2349_{\pm0.0004}$ \\
 & LUNA-Base & $28.86_{\pm0.50}$ & $55.71_{\pm1.37}$ & $51.67_{\pm0.54}$ & $78.12_{\pm0.61}$ & $49.96_{\pm0.35}$ & $50.00_{\pm0.00}$ & $0.2342_{\pm0.0043}$ \\
 & MIRepNet & $\underline{54.21}_{\pm0.24}$ & $69.57_{\pm0.54}$ & -- & -- & -- & -- & -- \\
 & \textbf{STEAM} & $53.43_{\pm0.27}$ & $\underline{72.26}_{\pm0.49}$ & $\mathbf{78.63}_{\pm0.54}$ & $\mathbf{81.75}_{\pm0.53}$ & $\underline{57.20}_{\pm0.30}$ & $\mathbf{77.90}_{\pm0.49}$ & $\mathbf{0.2281}_{\pm0.0036}$ \\
 & \textbf{STEAM-Spec} & $\mathbf{55.29}_{\pm0.46}$ & $\mathbf{73.39}_{\pm0.14}$ & -- & -- & $\mathbf{60.83}_{\pm0.30}$ & -- & -- \\
\midrule
\multicolumn{9}{l}{\textit{Within-subject (Few-shot)}} \\
\multirow{6}{*}{\begin{tabular}{@{}c@{}}Specialist\\models\end{tabular}} & EEGNet & $50.22_{\pm1.14}$ & $72.08_{\pm0.39}$ & $\mathbf{68.99}_{\pm0.51}$ & $88.45_{\pm0.48}$ & $52.12_{\pm0.47}$ & $60.57_{\pm1.29}$ & $0.2082_{\pm0.0045}$ \\
 & ShallowConv & $52.87_{\pm0.88}$ & $73.79_{\pm0.66}$ & $57.13_{\pm0.93}$ & $85.81_{\pm0.44}$ & $51.97_{\pm0.26}$ & $69.37_{\pm1.43}$ & $0.3839_{\pm0.0063}$ \\
 & LMDA & $51.13_{\pm0.76}$ & $73.79_{\pm1.50}$ & $60.28_{\pm0.82}$ & $86.80_{\pm0.55}$ & $53.20_{\pm1.39}$ & $54.78_{\pm0.22}$ & $0.2027_{\pm0.0110}$ \\
 & CNN-T & $51.27_{\pm1.10}$ & $71.63_{\pm1.36}$ & $50.32_{\pm0.52}$ & $87.66_{\pm0.33}$ & $51.95_{\pm1.06}$ & $51.08_{\pm0.95}$ & $\underline{0.1538}_{\pm0.0100}$ \\
 & Deformer & $42.21_{\pm0.73}$ & $70.32_{\pm1.28}$ & $65.38_{\pm0.32}$ & $86.88_{\pm0.33}$ & $52.19_{\pm0.30}$ & $52.01_{\pm0.39}$ & $0.2902_{\pm0.0167}$ \\
 & Conformer & $57.19_{\pm1.32}$ & $77.10_{\pm1.49}$ & $53.38_{\pm0.31}$ & $\mathbf{91.58}_{\pm0.34}$ & $55.67_{\pm1.63}$ & $68.33_{\pm1.71}$ & $\mathbf{0.1421}_{\pm0.0034}$ \\
\cmidrule(lr){1-9}
\multirow{13}{*}{\begin{tabular}{@{}c@{}}Foundation\\models\end{tabular}} & BENDR & $44.90_{\pm1.31}$ & $56.59_{\pm0.25}$ & $59.27_{\pm1.03}$ & $54.14_{\pm0.33}$ & $41.03_{\pm0.54}$ & $52.16_{\pm0.22}$ & $0.2436_{\pm0.0023}$ \\
 & BIOT & $48.60_{\pm0.72}$ & $68.43_{\pm1.48}$ & $52.19_{\pm0.16}$ & $79.60_{\pm0.51}$ & $48.41_{\pm0.56}$ & $\underline{86.11}_{\pm3.16}$ & $0.2230_{\pm0.0414}$ \\
 & LaBraM & $37.13_{\pm0.92}$ & $61.39_{\pm0.65}$ & $60.06_{\pm0.58}$ & $71.31_{\pm0.29}$ & $47.00_{\pm0.92}$ & $65.74_{\pm1.50}$ & $0.1956_{\pm0.0017}$ \\
 & Neuro-GPT & $42.18_{\pm0.23}$ & $61.90_{\pm0.90}$ & $55.02_{\pm0.97}$ & $83.09_{\pm0.59}$ & ${55.90}_{\pm0.09}$ & $71.22_{\pm3.69}$ & $0.1880_{\pm0.0035}$ \\
 & EEGPT & $34.99_{\pm0.25}$ & $59.17_{\pm0.87}$ & $52.02_{\pm1.14}$ & $63.74_{\pm0.28}$ & $40.48_{\pm0.44}$ & $65.59_{\pm2.51}$ & $0.1990_{\pm0.0007}$ \\
 & CBraMod & $50.34_{\pm1.18}$ & $70.30_{\pm0.72}$ & $56.85_{\pm1.03}$ & $\underline{88.54}_{\pm0.16}$ & $55.82_{\pm0.39}$ & $79.32_{\pm1.15}$ & $0.2051_{\pm0.0036}$ \\
 & TFM & $33.30_{\pm0.46}$ & $55.34_{\pm0.28}$ & $51.42_{\pm0.87}$ & $59.55_{\pm0.57}$ & $35.40_{\pm0.15}$ & $77.55_{\pm0.95}$ & $0.2208_{\pm0.0058}$ \\
 & BrainOmni & $39.00_{\pm0.19}$ & $61.59_{\pm1.54}$ & $57.02_{\pm0.36}$ & -- & $46.74_{\pm0.39}$ & $58.72_{\pm0.79}$ & $0.2146_{\pm0.0089}$ \\
 & SingLEM & $28.94_{\pm0.73}$ & $52.12_{\pm0.63}$ & $57.70_{\pm0.41}$ & $56.70_{\pm1.49}$ & $47.01_{\pm2.01}$ & $50.46_{\pm0.19}$ & $0.2271_{\pm0.0021}$ \\
 & LUNA-Base & $31.94_{\pm1.24}$ & $60.36_{\pm1.69}$ & $51.66_{\pm0.99}$ & $72.30_{\pm0.26}$ & $46.06_{\pm0.68}$ & $50.62_{\pm0.44}$ & $0.1676_{\pm0.0034}$ \\
 & MIRepNet & $\underline{63.27}_{\pm0.47}$ & $80.81_{\pm0.41}$ & -- & -- & -- & -- & -- \\
 & \textbf{STEAM} & $59.66_{\pm1.74}$ & $\underline{81.79}_{\pm0.21}$ & $\underline{68.19}_{\pm0.74}$ & $88.44_{\pm0.28}$ & $\underline{61.21}_{\pm0.35}$ & $\mathbf{92.90}_{\pm0.97}$ & $0.1625_{\pm0.0079}$ \\
 & \textbf{STEAM-Spec} & $\mathbf{65.47}_{\pm0.67}$ & $\mathbf{82.86}_{\pm0.43}$ & -- & -- &$\mathbf{65.18}_{\pm0.07}$ & -- & -- \\
\bottomrule
\end{tabular}
\endgroup
\caption{Performance on the seven datasets. Classification datasets report BAC (\%), and SEED-VIG reports RMSE. The best results are marked in bold, and the second-best results are underlined. STEAM-Spec denotes the corresponding MI- or emotion-specialized variant. `--' indicates that the corresponding scenario is not applicable.}
\label{tab:bench_selected}
\end{table*}

\subsection{Experiment Settings}

EEG signals are resampled and segmented using dataset-specific windows. Stage-I optimizes masked reconstruction and spatio-temporal contrastive learning. Stage-II adds paradigm-specific supervision to obtain STEAM-Spec variants. During downstream adaptation, all models use full fine-tuning; few-shot calibration uses limited target-subject trials. Balanced accuracy (BAC) evaluates classification, whereas root mean square error (RMSE) evaluates SEED-VIG regression. Results are averaged over three runs.

Specialist baselines comprise EEGNet~\cite{lawhern2018eegnet}, ShallowConvNet~\cite{schirrmeister2017deep}, LMDA-Net~\cite{miao2023lmdanet}, CNN-T~\cite{peh2022transformer}, EEG-Deformer~\cite{ding2024eeg}, and EEG-Conformer~\cite{song2022eeg}. Foundation-model baselines comprise BENDR~\cite{kostas2021bendr}, BIOT~\cite{yang2023biot}, LaBraM~\cite{jiang2024labram}, Neuro-GPT~\cite{cui2024neurogpt}, EEGPT~\cite{wang2024eegpt}, CBraMod~\cite{wang2025cbramod}, TFM-Tokenizer~\cite{pradeepkumar2025tokenizing}, BrainOmni~\cite{xiao2026brainomni}, SingLEM~\cite{sukhbaatar2025singlem}, LUNA-Base~\cite{doner2026luna}, and MIRepNet~\cite{liu2026mirepnet}.

\begin{table}[!t]
\centering
\begingroup
\small
\setlength{\tabcolsep}{1.2pt}
\renewcommand{\arraystretch}{1.06}
\begin{tabular*}{\columnwidth}{@{\extracolsep{\fill}}lccccccc@{}}
\toprule
\raisebox{0.5\normalbaselineskip}{Variant} & \raisebox{0.5\normalbaselineskip}{Spa.} & \raisebox{0.5\normalbaselineskip}{Temp.} & \raisebox{0.5\normalbaselineskip}{Gate} & \raisebox{0.5\normalbaselineskip}{SSMoE} & \shortstack{BNCI\\2015001} & \raisebox{0.5\normalbaselineskip}{CHB-MIT} & \raisebox{0.5\normalbaselineskip}{SEED} \\
\midrule
Full & $\checkmark$ & $\checkmark$ & $\checkmark$ & $\checkmark$ & \textbf{81.79} & \textbf{88.44} & \textbf{61.21} \\
w/o Spa. & $\times$ & $\checkmark$ & $\checkmark$ & $\checkmark$ & 80.06 & 87.72 & 60.56 \\
w/o Temp. & $\checkmark$ & $\times$ & $\checkmark$ & $\checkmark$ & 58.33 & 52.02 & 47.09 \\
w/o Gate & $\checkmark$ & $\checkmark$ & $\times$ & $\checkmark$ & 69.82 & 88.08 & 58.15 \\
w/o SSMoE & $\checkmark$ & $\checkmark$ & $\checkmark$ & $\times$ & 78.39 & 87.41 & 59.90 \\
\bottomrule
\end{tabular*}
\endgroup
\caption{Component ablations under within-subject few-shot evaluation. Values are BAC (\%).}
\label{tab:component_ablation}
\end{table}

\subsection{Main Results}

We evaluated all methods under cross-subject leave-one-subject-out (LOSO) and within-subject few-shot protocols using limited target-subject labels. Fig.~\ref{fig:fm_cost_rank} summarizes average rank, inference FLOPs, and parameter count for STEAM and the compared EEG foundation models. Stage-I STEAM achieves the best average rank with low inference FLOPs, indicating a favorable performance--efficiency trade-off.

As shown in Table~\ref{tab:bench_selected}, under cross-subject evaluation, STEAM shows strong transfer to non-MI tasks, achieving the best or tied-best performance on BNCI2014009, CHB-MIT, SEED, EEGMAT, and SEED-VIG. Under within-subject few-shot evaluation, STEAM remains competitive across tasks.   STEAM-Spec consistently improves upon Stage-I STEAM on BNCI2014001, BNCI2015001, and SEED under both protocols, achieving the best performance among the compared methods. These results demonstrate broad Stage-I transferability and effective Stage-II paradigm specialization. Additional results appear in Appendix~C.

\subsection{Model-component Ablations}

Table~\ref{tab:component_ablation} evaluates each architectural component under within-subject few-shot evaluation. Removing either branch degrades performance, confirming the complementary roles of spatial and temporal representations, while the frequency-aware gate consistently benefits EEG decoding. Notably, bypassing SSMoE leads to performance drops across datasets, demonstrating that independent branch encoding is insufficient. This validates SSMoE for effective cross-branch interaction and high-level spatio-temporal integration.

\subsection{Effectiveness of Hierarchical Pre-training and Lightweight Adaptation}

Table~\ref{tab:pretraining_stage_ablation} evaluates hierarchical pre-training by comparing no pre-training, each stage alone, and the complete pipeline under the same downstream protocol. Each stage independently improves downstream performance, while their combination achieves the best results, confirming the complementary benefits of general representation learning and paradigm-specific specialization.

\begin{table}[!t]
    \centering
    \begingroup
    \small
    \setlength{\tabcolsep}{2.2pt}
    \renewcommand{\arraystretch}{1.06}
    \begin{tabular*}{\columnwidth}{@{\extracolsep{\fill}}lcccc@{}}
        \toprule
        Setting & Stage-I & Stage-II & BNCI2015001 & SEED \\
        \midrule
        w/o Pre-training  & $\times$     & $\times$     & 79.34          & 57.83          \\
        Stage-I only     & $\checkmark$ & $\times$     & 81.79          & 61.21          \\
        Stage-II only    & $\times$     & $\checkmark$ & 81.93          & 63.03          \\
        Full Pre-training & $\checkmark$ & $\checkmark$ & \textbf{82.86} & \textbf{65.18} \\
        \bottomrule
    \end{tabular*}
    \endgroup
    \caption{Ablation of the two-stage hierarchical pre-training strategy. Values are BAC (\%).}
    \label{tab:pretraining_stage_ablation}
\end{table}

\begin{figure}[!t]
    \centering
    \includegraphics[width=\columnwidth]{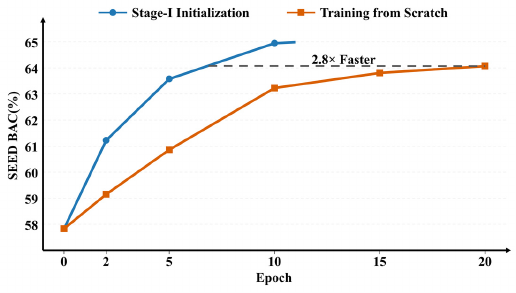}
    \caption{Effect of Stage-I initialization on Stage-II paradigm specialization.
    Each point reports the downstream SEED BAC obtained from the checkpoint at the corresponding Stage-II epoch.
    Stage-I initialization reaches the marked accuracy $2.8\times$ faster.}
    \label{fig:stage2_initialization}
\end{figure}

Figure~\ref{fig:stage2_initialization} further examines how Stage-I initialization facilitates Stage-II specialization. Compared with training from scratch, Stage-I initialization yields stronger downstream performance throughout training and reaches the marked SEED accuracy in $2.8\times$ fewer epochs, demonstrating more effective and efficient paradigm-specific pre-training.

We further evaluated lightweight adaptation by freezing the Transformer backbone and updating only the spatial and temporal embedding modules and task head, which account for approximately $5.1\%$ of the parameters. As shown in Table~\ref{tab:pretrain_peft_vs_full}, this strategy remains competitive with full fine-tuning on most classification tasks but is less effective for regression.

\begin{table}[!t]
    \centering
    \begingroup
    \small
    \setlength{\tabcolsep}{2pt}
    \renewcommand{\arraystretch}{1.06}
        \begin{tabular*}{\columnwidth}{@{\extracolsep{\fill}}lccccc@{}}
            \toprule
            Dataset & Metric & PEFT & Full FT & Gap & PEFT Params. \\
            \midrule
            BNCI2015001 & BAC  & 83.93  & 82.86  & $+1.07$   & \multirow{7}{*}{$\sim$5.1\%} \\
            BNCI2014001 & BAC  & 64.22  & 65.47  & $-1.25$   &                              \\
            BNCI2014009 & BAC  & 67.89  & 68.20  & $-0.31$   &                              \\
            SEED        & BAC  & 64.64  & 65.18  & $-0.54$   &                              \\
            CHB-MIT     & BAC  & 89.84  & 88.44  & $+1.40$   &                              \\
            EEGMAT      & BAC  & 94.91  & 92.90  & $+2.01$   &                              \\
            SEED-VIG    & RMSE & 0.2037 & 0.1625 & $+0.0408$ &                              \\
            \bottomrule
        \end{tabular*}
    \endgroup
    \caption{Comparison of lightweight adaptation and full fine-tuning.
    BAC is reported in percentage points, while SEED-VIG reports RMSE.}
    \label{tab:pretrain_peft_vs_full}
\end{table}

\subsection{Analysis of SSMoE Routing}

\begin{figure}[!t]
\centering
\includegraphics[width=\columnwidth]{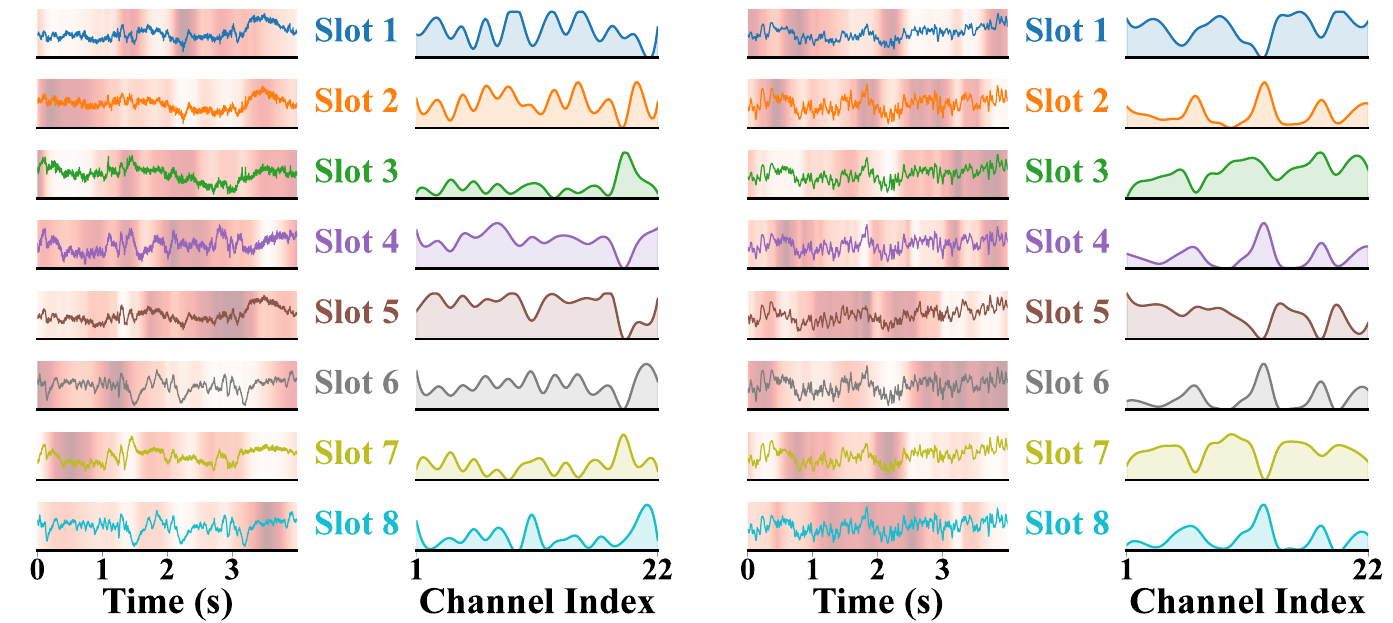}
\caption{Slot-wise SSMoE visualization on TUH (left) and BNCI2014001 (right). Each visualization shows the temporal focus projected onto EEG traces and the corresponding channel-attention profile.}
\label{fig:ssmoe_slot_analysis}
\end{figure}

To examine whether the SSMoE slots exhibit differentiated routing behavior, we visualize their slot-wise responses for representative trials, as shown in Fig.~\ref{fig:ssmoe_slot_analysis}. Each visualization projects the temporal routing response of each slot onto the corresponding EEG trace and presents its channel-wise routing profile. Different slots emphasize distinct temporal intervals and channel subsets, suggesting that they do not collapse to an identical global routing pattern. This qualitative evidence is consistent with the intended role of SSMoE as a compact communication interface between the spatial and temporal branches, through which EEG information can be aggregated and redistributed. 

\section{Conclusion}

We presented STEAM, an EEG foundation model that bridges broad cross-paradigm transfer and paradigm-specific specialization. STEAM combines a dual-branch spatio-temporal encoder, shared SSMoE-based cross-branch interaction, and frequency-aware gated attention to capture complementary spatial, temporal, and spectral structures. The general-to-paradigm hierarchical pre-training strategy progressively specializes a reusable backbone for a target paradigm without training a separate model from scratch. Experiments on seven datasets demonstrated strong transferability and computational efficiency, while Stage-II pre-training consistently improved paradigm-specific decoding. Lightweight adaptation also remained competitive with full fine-tuning on most classification tasks, further supporting the transferability and reusability of the learned representations.

\bibliography{aaai2027}

\clearpage
\appendix
\setcounter{secnumdepth}{1}

\section{Model Architecture, Computational Cost, and Reproducibility}
\label{app:architecture}

\subsection{Detailed Architecture}

The architectural and FLOP analyses were conducted using a four-second EEG segment sampled at 250~Hz as the reference input, $\mathbf{X}\in\mathbb{R}^{60\times1000}$. STEAM processes the input through parallel spatial and temporal pathways to preserve complementary channel-wise and temporal structures before cross-branch interaction. The spatial and temporal tokenizers produce 60 and 61 tokens, respectively, with an embedding dimension of 128. The resulting token sequences are processed by two separate six-layer Transformer encoders. The implementation also accommodates the dataset-specific input durations described in Section~B, for which only the number of temporal tokens varies. Table~\ref{tab:architecture} summarizes the layer-wise tensor shapes for the reference input, where $B$ denotes the batch size and $K$ denotes the number of task classes.

\begin{table*}[t]
\centering
\begingroup
\small
\setlength{\tabcolsep}{1mm}
\renewcommand{\arraystretch}{1.3}
\begin{tabular*}{\textwidth}{@{\extracolsep{\fill}}lcccc@{}}
\toprule
Branch & Operation & Configuration & Input shape & Output shape \\
\midrule
Input
& EEG segment
& 250~Hz; 4~s
& $B \times 60 \times 1000$
& $B \times 60 \times 1000$ \\

Spatial
& Channel-wise reshape
& ---
& $B \times 60 \times 1000$
& $(B \cdot 60) \times 1 \times 1000$ \\

Spatial
& Conv1d
& $1 \!\rightarrow\! 40$, $k=25$, $s=5$, $p=12$
& $(B \cdot 60) \times 1 \times 1000$
& $(B \cdot 60) \times 40 \times 200$ \\

Spatial
& Adaptive average pooling
& output size $=1$
& $(B \cdot 60) \times 40 \times 200$
& $(B \cdot 60) \times 40 \times 1$ \\

Spatial
& Linear projection
& $40 \!\rightarrow\! 128$
& $B \times 60 \times 40$
& $B \times 60 \times 128$ \\

Temporal
& Dimension expansion
& ---
& $B \times 60 \times 1000$
& $B \times 1 \times 60 \times 1000$ \\

Temporal
& Temporal Conv2d
& $1 \!\rightarrow\! 64$, $k=(1,25)$
& $B \times 1 \times 60 \times 1000$
& $B \times 64 \times 60 \times 976$ \\

Temporal
& Spatial Conv2d
& $64 \!\rightarrow\! 128$, $k=(60,1)$
& $B \times 64 \times 60 \times 976$
& $B \times 128 \times 1 \times 976$ \\

Temporal
& Average pooling
& $k=(1,75)$, $s=(1,15)$
& $B \times 128 \times 1 \times 976$
& $B \times 128 \times 1 \times 61$ \\

Temporal
& $1 \times 1$ Conv2d
& $128 \!\rightarrow\! 128$
& $B \times 128 \times 1 \times 61$
& $B \times 128 \times 1 \times 61$ \\

Temporal
& Token-sequence reshape
& ---
& $B \times 128 \times 1 \times 61$
& $B \times 61 \times 128$ \\

Spatial
& Transformer blocks 
& 8 attention heads; $FFN dim = 128$
& $B \times N_s \times 128$
& $B \times N_s \times 128$ \\

Temporal
& Transformer blocks 
& 8 attention heads; $FFN dim = 128$
& $B \times N_t \times 128$
& $B \times N_t \times 128$ \\

Both branches
& SSMoE blocks 
& 8 experts; 8 slots
& $B \times (N_s+N_t) \times 128$
& $B \times (N_s+N_t) \times 128$ \\

Both branches
& Reconstruction decoder
& 2 Transformer blocks
& $B \times N_q \times 128$
& $B \times N_q \times 128$ \\

Downstream
& Classification head
& $128 \!\rightarrow\! K$
& $B \times 128$
& $B \times K$ \\
\bottomrule
\end{tabular*}
\endgroup
\caption{
Layer-wise tensor shapes of STEAM for a four-second reference input.
Here, $B$ denotes the batch size, $K$ denotes the number of task classes,
and $N_s=60$ and $N_t=61$ denote the numbers of spatial and temporal
tokens, respectively.
For branch $q \in \{s,t\}$, $N_q$ denotes the corresponding branch-specific
token count.
The symbols $k$, $s$, and $p$ denote kernel size, stride, and padding,
respectively.
}
\label{tab:architecture}
\end{table*}

Frequency-aware gates modulate the branch representations at each encoder layer. A shared SSMoE module then operates on the concatenated spatial and temporal tokens, using eight compact slots to aggregate and redistribute cross-branch information. In our implementation, each expert is paired with exactly one slot, establishing a one-to-one correspondence between experts and slots. The resulting sequence is subsequently divided into the corresponding spatial and temporal pathways. During downstream adaptation, the model is initialized with the pre-trained embedding modules and branch encoders. Full fine-tuning updates all model parameters, whereas Pretrain-PEFT updates only the spatial and temporal embedding modules and the task-specific head. The two embedding modules contain 0.52 million parameters, corresponding to approximately 5.1\% of the Stage-I model parameters, excluding the downstream task head.

\subsection{Module-wise Parameter Count}

Table~\ref{tab:parameters} summarizes the parameter distribution across the principal components of STEAM. Because Stage-I and Stage-II use the same backbone architecture, both models contain approximately 10.23 million parameters. Dataset-specific classification heads and auxiliary modules used only during optimization are excluded from the reported totals; the optimization-specific auxiliary modules contain $6.60\times10^{4}$ parameters. Removing the mask tokens and reconstruction decoders, which are unnecessary for downstream inference, reduces the model to a 9.43-million-parameter inference encoder. 

\begin{table*}[t]
\centering
\begingroup
\small
\setlength{\tabcolsep}{1mm}
\renewcommand{\arraystretch}{1.3}
\begin{tabular*}{\textwidth}{@{\extracolsep{\fill}}lcc@{}}
\toprule
Module & Parameters (M) & Share of Stage-I total (\%) \\
\midrule
Spatial tokenizer
& $6.25\times10^{-3}$
& 0.06 \\

Temporal tokenizer
& 0.51
& 4.99 \\

Spatial Transformer stack
& 1.19
& 11.63 \\

Temporal Transformer stack
& 1.19
& 11.63 \\

SSMoE stack
& 6.33
& 61.88 \\

Spatial frequency-aware gates
& 0.10
& 1.02 \\

Temporal frequency-aware gates
& 0.10
& 1.02 \\

Spatial and temporal mask tokens
& $2.56\times10^{-4}$
& $<0.01$ \\

Spatial reconstruction decoder
& 0.40
& 3.88 \\

Temporal reconstruction decoder
& 0.40
& 3.88 \\
\midrule

Stage-I total
& 10.23
& 100.00 \\

Stage-II total
& 10.23
& --- \\

Inference encoder
& 9.43
& --- \\
\bottomrule
\end{tabular*}
\endgroup
\caption{
Module-wise parameter count of STEAM.
Parameter counts are reported in millions.
Values below 0.01 million are shown in scientific notation,
while all other values are rounded to two decimal places.
The reported shares are computed relative to the complete
Stage-I model and may not sum to exactly 100\% because of
rounding.
Dataset-specific classification heads are excluded.
}
\label{tab:parameters}
\end{table*}

Although SSMoE constitutes a substantial proportion of the model parameters, its inference overhead remains limited because cross-branch communication is performed through only eight compact slots.

\subsection{Stage-I Pre-training Configuration}

Stage-I jointly optimized masked reconstruction, token-diversity regularization, and cross-view contrastive alignment. The reconstruction loss was computed only at masked positions, while the auxiliary objectives encouraged diverse token representations and semantic consistency between the spatial and temporal branches. Table~\ref{tab:stage1_config} summarizes the principal hyperparameters used to obtain the final Stage-I checkpoint.

\begin{table*}[t]
\centering
\begingroup
\small
\setlength{\tabcolsep}{1mm}
\renewcommand{\arraystretch}{1.3}
\begin{tabular*}{0.78\textwidth}{@{\extracolsep{\fill}}lc@{}}
\toprule
Hyperparameter & Value \\
\midrule
Training epochs & 5 \\
Training / validation split (\%) & 90 / 10 \\
Batch size & 1,024 \\
Optimizer & AdamW \\
Initial learning rate & $3\times10^{-4}$ \\
Learning-rate schedule & Per-iteration cosine decay \\
Weight decay & $1\times10^{-4}$ \\
Masking ratio & 0.50 \\
Diversity-loss weight ($\alpha$) & 0.30 \\
InfoNCE-loss weight ($\lambda$) & 0.01 \\
InfoNCE temperature ($\tau$) & 0.20 \\
Diversity-loss temperature & 1.00 \\
Contrastive-loss warm-up epoch & 1 \\
\bottomrule
\end{tabular*}
\endgroup
\caption{Principal optimization hyperparameters for Stage-I pre-training.}
\label{tab:stage1_config}
\end{table*}

\subsection{Stage-II Pre-training Configuration}

Stage-II continued pre-training from the final Stage-I checkpoint on multiple datasets within a target paradigm and introduced an additional supervised classification objective. The Stage-I objectives and their weights were retained, while the reconstruction and classification terms were assigned unit weights. This continual pre-training strategy preserved the general representations learned in Stage-I while adapting the shared backbone for paradigm-specific discrimination. Table~\ref{tab:stage2_config} summarizes the principal optimization hyperparameters.

\begin{table*}[t]
\centering
\begingroup
\small
\setlength{\tabcolsep}{1mm}
\renewcommand{\arraystretch}{1.3}
\begin{tabular*}{0.78\textwidth}{@{\extracolsep{\fill}}lc@{}}
\toprule
Hyperparameter & Value \\
\midrule
Training epochs & 20 \\
Training / validation split (\%) & 90 / 10 \\
Batch size & 256 \\
Optimizer & AdamW \\
Initial learning rate & $1\times10^{-4}$ \\
Learning-rate schedule & Per-iteration cosine decay \\
Weight decay & $1\times10^{-4}$ \\
Masking ratio & 0.50 \\
Classification-loss weight ($\mu$) & 1.00 \\
Diversity-loss weight ($\alpha$) & 0.30 \\
InfoNCE-loss weight ($\lambda$) & 0.01 \\
InfoNCE temperature ($\tau$) & 0.20 \\
Diversity-loss temperature & 1.00 \\
Contrastive-loss warm-up & 1 epoch \\
\bottomrule
\end{tabular*}
\endgroup
\caption{Principal optimization hyperparameters for Stage-II paradigm-specific continual pre-training.}
\label{tab:stage2_config}
\end{table*}

\subsection{Implementation Configuration}

The reported pre-training experiments were conducted on a single NVIDIA A100-SXM4 GPU with 80~GB of memory. The implementation used Python 3.10.19, PyTorch 2.6.0, and CUDA 12.4.

\section{Pre-training and Downstream Datasets}
\label{app:datasets}

\subsection{Pre-training and Evaluation Data Partitioning}

The datasets were partitioned into non-overlapping corpora for hierarchical pre-training and downstream evaluation. TUH EEG was used exclusively for Stage-I general pre-training. BNCI2014002, Cho2017, Dreyer2023, Lee2019, Weibo2014, Zhou2016, and PhysioNetMI formed the Stage-II motor-imagery corpus, while SEED-V and DEAP formed a separate Stage-II emotion-recognition corpus. BNCI2014001, BNCI2015001, BNCI2014009, CHB-MIT, EEGMAT, SEED, and SEED-VIG were reserved exclusively for downstream evaluation and excluded from both pre-training stages. Consequently, neither downstream MI dataset contributed to MI-specific pre-training, and SEED was not used for emotion-specific pre-training. A newly initialized dataset-specific prediction head was used for each downstream task.

\begin{table*}[t]
\centering
\begingroup
\small
\setlength{\tabcolsep}{1mm}
\renewcommand{\arraystretch}{1.3}
\noindent\textbf{(a) Stage-I general EEG corpus}\par\smallskip
\begin{tabular*}{\textwidth}{@{\extracolsep{\fill}}lccccccc@{}}
\toprule
Dataset & Recordings & Segments & Hz & Loaded ch. & Model ch. & Window (s) & Train / val. (\%) \\
\midrule
TUH EEG & 6,451 & 1,640,700 & 250 & 20 & 60 & 4.00 & 90.00 / 10.00 \\
\bottomrule
\end{tabular*}\par\medskip

\noindent\textbf{(b) Stage-II motor-imagery corpus}\par\smallskip
\begin{tabular*}{\textwidth}{@{\extracolsep{\fill}}lccccc@{}}
\toprule
Dataset & Trials (raw/used) & Ch. & Hz & Window (s) & Class distribution \\
\midrule
BNCI2014002 & 2,240 / 1,400 & 15 & 512 & 4.00 & Feet 700; right 700 \\
Cho2017 & 10,520 / 6,200 & 63 & 512 & 4.00 & Left 3,120; right 3,080 \\
Dreyer2023 & 16,152 / 16,152 & 27 & 512 & 4.00 & Left 8,074; right 8,078 \\
Lee2019 & 10,800 / 10,800 & 62 & 1,000 & 4.00 & Left 5,400; right 5,400 \\
Weibo2014 & 5,540 / 2,370 & 60 & 200 & 4.00 & Feet / left / right 790 each \\
Zhou2016 & 1,800 / 900 & 14 & 250 & 4.00 & Feet 301; left 300; right 299 \\
PhysioNetMI & 19,676 / 7,373 & 61 & 160 & 4.00 & Feet 2,455; left 2,480; right 2,438 \\
\midrule
Total & 66,728 / 45,195 & --- & --- & 4.00 & Feet 4,246; left 20,164; right 20,785 \\
\bottomrule
\end{tabular*}\par\medskip

\noindent\textbf{(c) Stage-II emotion corpus}\par\smallskip
\begin{tabular*}{\textwidth}{@{\extracolsep{\fill}}lccccc@{}}
\toprule
Dataset & Segments & Hz & Window (s) & Classes & Retained composition \\
\midrule
SEED-V & 24,768 & 1,000 & 1.00 & 3 & Neg. / neu. / pos.: 9,040 / 8,640 /  7,088 \\
DEAP & 76,800 & 128 & 1.00 & 2 & Neg. / pos.: 44,220 / 32,580 \\
\midrule
Total & 101,568 & --- & 1.00 & 3 & Neg. / neu. / pos.: 53,260 / 8,640 / 39,668 \\
\bottomrule
\end{tabular*}
\endgroup
\caption{Stage-I pre-training: (a) the processed Stage-I general EEG corpus, (b) the Stage-II motor-imagery corpus, and (c) the Stage-II emotion-recognition corpus. Class counts are reported after dataset-specific selection and preprocessing. In (c), neg., neu., and pos. denote negative, neutral, and positive, respectively.}
\label{tab:pretraining_datasets}
\end{table*}

\begin{table*}[t]
\centering
\begingroup
\small
\setlength{\tabcolsep}{1mm}
\renewcommand{\arraystretch}{1.3}
\begin{tabular*}{\textwidth}{@{\extracolsep{\fill}}llccccclcc@{}}
\toprule
Dataset & Task & Subj. & $N$ & Ch. & Hz & Input / model (s) & Target composition & Metric & Few-shot (\%) \\
\midrule
BNCI2014001 & MI & 9 & 2,592 & 22 & 250 & 4.00 / 4.00 & 4 classes, 648/class & BAC & 30.00 \\
BNCI2015001 & MI & 12 & 2,400 & 13 & 512 & 5.00 / 4.00 & 2 classes, 1,200/class & BAC & 30.00 \\
BNCI2014009 & P300 & 10 & 5,760 & 16 & 256 & 0.80 / 1.00 & Target/non-target: 960/4,800 & BAC & 10.00 \\
CHB-MIT & Seizure & 23 & 29,840 & 18 & 256 & 4.00 / 4.00 & Ictal/interictal: 2,690/27,150 & BAC & $\sim$12.75 \\
EEGMAT & Cognition & 36 & 1,080 & 19 & 500 & 4.00 / 4.00 & 2 classes, 540/class & BAC & 60.00 \\
SEED & Emotion & 15 & 50,910 & 62 & 200 & 1.00 / 1.00 & Neg. / neu. / pos.: 16,800 / 16,560 / 17,550 & BAC & $\sim$20.00 \\
SEED-VIG & Vigilance & 21 & 18,585 & 17 & 200 & 8.00 / 8.00 & Range: $[0.00,1.00]$ & RMSE & 10.00 \\
\bottomrule
\end{tabular*}
\endgroup
\caption{Downstream benchmark datasets and few-shot protocols. $N$ denotes the number of retained examples. MI denotes motor imagery; neg., neu., and pos. denote negative, neutral, and positive, respectively.}
\label{tab:downstream_datasets}
\end{table*}

\subsection{Stage-I General EEG Corpus}

\paragraph{TUH EEG.} TUH EEG provided heterogeneous clinical recordings for learning paradigm-general EEG representations. The preprocessing pipeline constructed an ordered 20-channel matrix, filled missing channels with zeros, applied common-average referencing, rescaled signal amplitudes by $10^3$, and performed resampling followed by non-overlapping segmentation. No additional band-pass filtering was applied during the final Stage-I preprocessing.

\subsection{Stage-II Paradigm-Specific Corpora}

\paragraph{Motor imagery.} The Stage-II MI corpus combined seven datasets within a shared label space comprising feet, left-hand, and right-hand imagery. Following dataset-specific trial selection and channel processing, all datasets were filtered between 0.5 and 40~Hz using a fifth-order filter, resampled to 250~Hz, and Euclidean-aligned within the selected native channel space of each dataset. The signals were then mapped to a standard 60-channel montage through inverse-distance interpolation, and each trial was cropped or zero-padded to $1{,}000$ samples.

\paragraph{BNCI2014002.} BNCI2014002 samples were restricted to feet- and right-hand-imagery trials, yielding a balanced two-class subset of the shared MI label space. All participant blocks were retained.

\paragraph{Cho2017.} Cho2017 provided high-density left- and right-hand-imagery recordings with a nearly balanced class distribution.

\paragraph{Dreyer2023.} Dreyer2023 provided a large and nearly balanced collection of left- and right-hand-imagery trials. All available participant blocks and trials were retained after channel processing.

\paragraph{Lee2019.} Lee2019 provided high-density left- and right-hand-imagery recordings with a balanced class distribution. All participant blocks and trials were retained.

\paragraph{Weibo2014.} Weibo2014 samples were restricted to feet-, left-hand-, and right-hand-imagery trials. Combined-limb imagery, generic hand imagery, and rest trials from the original seven-class formulation were excluded, yielding a balanced three-class subset of the shared MI label space.

\paragraph{Zhou2016.} Zhou2016 provided feet-, left-hand-, and right-hand-imagery trials with an approximately balanced class distribution.

\paragraph{PhysioNetMI.} PhysioNetMI samples were restricted to left-hand-, right-hand-, and feet-imagery trials through label-based filtering. All available participant blocks were retained.

\paragraph{Emotion recognition.} The Stage-II emotion corpus combined SEED-V and DEAP within a shared semantic label space comprising negative, neutral, and positive emotions. SEED-V and the downstream SEED dataset contained no overlapping participants. Both datasets were filtered between 0.5 and 40~Hz using a fifth-order filter, resampled to 250~Hz, aligned within each source participant, interpolated to the standard 60-channel montage, and segmented into one-second windows.

\paragraph{SEED-V.} SEED-V samples were restricted to the negative, neutral, and positive categories in the shared emotion label space. Signal amplitudes were converted from volts to microvolts before applying the common preprocessing pipeline.

\paragraph{DEAP.} DEAP contributed negative and positive samples to the shared emotion corpus. Its binary emotion labels were mapped to the corresponding negative and positive categories, and no neutral samples were introduced.

\subsection{Downstream Benchmark Datasets}

The downstream benchmark comprised seven datasets spanning six representative EEG applications: motor imagery (MI), event-related potential detection, seizure detection, cognitive-state recognition, emotion recognition, and continuous vigilance estimation. Table~\ref{tab:downstream_datasets} summarizes the signal characteristics, target distributions, evaluation metrics, and labeled-data fractions used in the few-shot setting. Dataset-specific selection and preprocessing procedures are described below. All downstream recordings were resampled to 250~Hz and mapped to the 60-channel model interface before segmentation.

\paragraph{BNCI2014001.} BNCI2014001 is a four-class cue-based MI dataset comprising left-hand, right-hand, feet, and tongue imagery. The designated training session was retained, and each trial was segmented into a four-second epoch. Performance was evaluated using balanced accuracy (BAC).

\paragraph{BNCI2015001.} BNCI2015001 contains right-hand and feet motor-imagery trials and was formulated as a binary classification task. The designated training session was retained, and the approximately five-second epochs were adjusted to the four-second input duration used by the MI model. Performance was evaluated using BAC.

\paragraph{BNCI2014009.} BNCI2014009 is a P300 oddball dataset formulated as binary target-versus-non-target classification. Epochs from the designated training session were mapped to a fixed 16-channel montage and adjusted to the one-second model input duration. BAC was used to account for the pronounced class imbalance.

\paragraph{CHB-MIT.}
CHB-MIT contains long-term pediatric scalp-EEG recordings and was formulated as a binary seizure-detection task. The recordings were converted to a common 18-channel bipolar montage, after which ictal and interictal segments were divided into four-second epochs. For within-subject evaluation, the first seizure event and the corresponding interictal period of each patient were used for adaptation, while the remaining seizure events and interictal data were reserved for testing. The split was performed before epoch segmentation, ensuring that epochs from the same seizure event or continuous recording period did not appear in both sets. Performance was evaluated using balanced accuracy (BAC).

\paragraph{EEGMAT.}
EEGMAT was used for binary cognitive-state recognition. For each participant, the final 60 seconds of the first condition and the initial 60 seconds of the second condition were retained. The within-subject split was performed at the continuous-block level before the retained signals were divided into non-overlapping four-second windows. Performance was evaluated using BAC.

\paragraph{SEED.}
SEED was used for three-class emotion recognition. The first recording session was retained. For within-subject evaluation, one complete continuous trial (video clip) from each emotion class was used for adaptation, while the remaining four complete trials from that class were reserved for testing. The trial-level split was performed before the signals were segmented into one-second windows, preventing temporally adjacent windows from being shared across the adaptation and test sets. Performance was evaluated using BAC.

\paragraph{SEED-VIG.} SEED-VIG was used for continuous vigilance estimation with PERCLOS-derived regression targets. Each participant contributed 885 eight-second EEG windows. Performance was evaluated using root mean square error (RMSE), with lower values indicating better predictive accuracy.

\subsection{Evaluation Protocols and Metrics}

Cross-subject evaluation follows a leave-one-subject-out protocol, in which all samples from the held-out participant are used exclusively for testing. This setting evaluates cross-subject transfer without using labeled data from the target participant. Within-subject few-shot evaluation instead adapts the model using the dataset-specific fractions of labeled target-participant data reported in Table~\ref{tab:downstream_datasets}. For each dataset and evaluation setting, all compared methods use identical data partitions.

For a $K$-class classification task, balanced accuracy (BAC) is defined as the unweighted mean of the per-class recalls:
\begin{equation}
\mathrm{BAC}=\frac{1}{K}\sum_{k=1}^{K}\frac{\mathrm{TP}_k}{\mathrm{TP}_k+\mathrm{FN}_k},
\label{eq:bac}
\end{equation}
where $\mathrm{TP}_k$ and $\mathrm{FN}_k$ denote the numbers of true-positive and false-negative samples for class $k$, respectively. BAC reduces the influence of class imbalance, particularly for tasks containing substantially more non-target P300 epochs or interictal EEG windows. For SEED-VIG, given the regression targets $y_i$ and predictions $\hat{y}_i$, performance is measured using root mean square error (RMSE):
\begin{equation}
\mathrm{RMSE}=\sqrt{\frac{1}{N}\sum_{i=1}^{N}(y_i-\hat{y}_i)^2},
\label{eq:rmse}
\end{equation}
where $N$ denotes the number of evaluation samples. Classification performance is reported as BAC (\%), whereas regression performance is reported on the normalized target scale. Unless otherwise specified, error bars in participant-level figures indicate the standard deviation across participants.

\section{Detailed Experimental Results and Ablation Studies}
\label{app:results}

\subsection{Effect of Pre-training and Adaptation Scope}

\begin{figure}[t]
\centering
\includegraphics[width=\columnwidth]{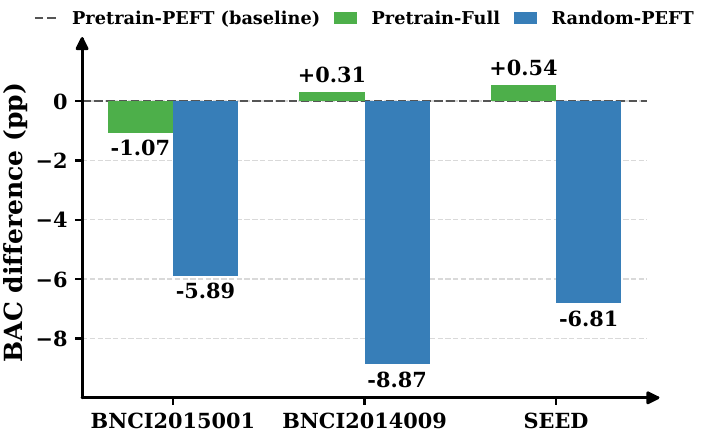}
\caption{BAC differences relative to Pretrain-PEFT under the few-shot protocol. Pretrain-Full fine-tunes all model parameters from the same pre-trained initialization, whereas Random-PEFT replaces the pre-trained Transformer backbone with a randomly initialized frozen backbone. Positive values indicate improvements over Pretrain-PEFT.}
\label{fig:pretrain_peft_gap}
\end{figure}

We further disentangled the contributions of backbone pre-training and downstream adaptation scope. Pretrain-PEFT initialized the encoder from pre-trained weights, froze the backbone, and updated only the spatial and temporal embedding modules and the newly initialized task head. Pretrain-Full used the same pre-trained initialization but updated all model parameters. Random-PEFT retained the pre-trained embedding initialization but replaced the pre-trained backbone with a randomly initialized frozen backbone, while updating the same lightweight components as Pretrain-PEFT. Figure~\ref{fig:pretrain_peft_gap} reports the BAC difference of each variant relative to Pretrain-PEFT.

Pretrain-Full remained close to Pretrain-PEFT across all three datasets and provided no consistent advantage, indicating that lightweight adaptation retained most of the benefits of full-model fine-tuning. By contrast, Random-PEFT incurred substantial performance degradation across all datasets, demonstrating the importance of the representations learned by the pre-trained backbone even when its parameters were frozen during downstream adaptation. These results indicated that the effectiveness of Pretrain-PEFT depended not only on the trainable embedding modules and task head, but also on reusing the transferable representations encoded by the pre-trained backbone.

\subsection{Pre-training Objective Ablations}

Table~\ref{tab:pretraining_ablation} evaluates the contributions of reconstruction and cross-view InfoNCE in Stage-I, together with reconstruction and supervised classification in Stage-II. Each ablation removed one objective while keeping the pre-training data, optimizer, learning-rate schedule, and training duration unchanged. The Stage-I variants were trained for five epochs and evaluated on BNCI2015001, CHB-MIT, and SEED. The Stage-II variants were trained for 20 epochs, with the MI- and emotion-specialized models evaluated on BNCI2015001 and SEED, respectively.

Removing either Stage-I objective degraded performance across all three downstream datasets, with reconstruction contributing particularly strongly to MI and emotion recognition. The complete Stage-II objective achieved the best performance on both target-paradigm datasets. Removing either reconstruction or supervised classification reduced performance, confirming that self-supervised representation preservation and paradigm-specific supervision provided complementary benefits during continual pre-training.

\begin{table*}[t]
\centering
\begingroup
\small
\renewcommand{\arraystretch}{1.3}
\setlength{\tabcolsep}{1mm}
\begin{tabular*}{\textwidth}{@{\extracolsep{\fill}}lccccccc@{}}
\toprule
Configuration & Stage-I Rec. & Stage-I InfoNCE & Stage-II Rec. & Stage-II Cls. & BNCI2015001 & CHB-MIT & SEED \\
\midrule
Full Stage-I & $\checkmark$ & $\checkmark$ & --- & --- & \textbf{81.79} & \textbf{88.44} & \textbf{61.21} \\
Stage-I w/o Rec. Loss & $\times$ & $\checkmark$ & --- & --- & 75.50 & 88.08 & 58.78 \\
Stage-I w/o InfoNCE Loss & $\checkmark$ & $\times$ & --- & --- & 80.71 & 87.22 & 60.31 \\
\midrule
Full Stage-II & $\checkmark$ & $\checkmark$ & $\checkmark$ & $\checkmark$ & \textbf{82.86} & --- & \textbf{65.18} \\
Stage-II w/o Rec. Loss & $\checkmark$ & $\checkmark$ & $\times$ & $\checkmark$ & 81.08 & --- & 63.51 \\
Stage-II w/o Cls. Loss & $\checkmark$ & $\checkmark$ & $\checkmark$ & $\times$ & 80.79 & --- & 62.55 \\
\bottomrule
\end{tabular*}
\endgroup
\caption{Pre-training objective ablations under the few-shot evaluation protocol. Each variant removes one objective while retaining the remaining objectives and the same training configuration. The Stage-I variants are evaluated on BNCI2015001, CHB-MIT, and SEED, whereas the Stage-II MI- and emotion-specialized variants are evaluated on BNCI2015001 and SEED, respectively. Rec. and Cls. denote reconstruction and supervised classification, respectively. Values are BAC (\%), and boldface indicates the best result within each pre-training stage on each dataset.}
\label{tab:pretraining_ablation}
\end{table*}

\subsection{Controlled Comparison of Cross-Branch Fusion}
\label{app}

To examine whether the gains of SSMoE stemmed from its shared expert routing rather than from generic cross-branch interaction, we compared Shared SSMoE with four independently trained structural alternatives. Bidirectional cross-attention employed separate spatial-to-temporal and temporal-to-spatial attention modules. MLP fusion concatenated each branch token with a pooled representation of the other branch and processed the resulting representation using a shared MLP. Late fusion removed intermediate token-level interaction and combined the pooled representations of the two branches only before the final projection. As a parameter-matched capacity control, Separate Soft-MoE assigned four experts independently to each branch, yielding eight experts in total and the same parameter count as Shared SSMoE.

Each alternative was pre-trained from scratch on the same Stage-I corpus for five epochs. The data partition, training objectives, optimizer, learning-rate schedule, batch size, model depth, and embedding dimension were held fixed. All models were subsequently evaluated using full-model adaptation under the same dataset-specific within-subject few-shot protocol.The w/o SSMoE experiment reported in the main text was not included as an MLP control because it bypassed intermediate cross-branch fusion in the original architecture. It therefore served as a runtime ablation rather than an independently trained fusion alternative.

\begin{table*}[t]
\centering
\begingroup
\small
\renewcommand{\arraystretch}{1.3}
\setlength{\tabcolsep}{2mm}
\begin{tabular*}{\textwidth}{@{\extracolsep{\fill}}lccc@{}}
\toprule
Fusion mechanism & Params (M) / FLOPs (G) & BNCI2015001 ($n=12$) & SEED ($n=15$) \\
\midrule
Late fusion & 3.93 / 1.48 & 78.17 & 57.58 \\
MLP fusion & 4.49 / 1.63 & 77.56 & 58.74 \\
Bidirectional cross-attention & 4.30 / 1.53 & 77.86 & 57.23 \\
Separate Soft-MoE & 10.23 / 1.49 & 80.42 & 56.54 \\
Shared SSMoE & 10.23 / 1.49 & \textbf{81.79} & \textbf{61.21} \\
\bottomrule
\end{tabular*}
\endgroup
\caption{Controlled comparison of cross-branch fusion mechanisms under the dataset-specific within-subject few-shot protocols. Values denote participant-level mean BAC (\%). Parameter counts represent the total number of model parameters rounded to millions, and boldface indicates the highest mean on each dataset.}
\label{tab}
\end{table*}

As shown in Table~\ref{tab}, Shared SSMoE achieved the best performance on both datasets, demonstrating the strongest cross-dataset consistency among all fusion mechanisms. In contrast, the parameter-matched Separate Soft-MoE performed competitively on BNCI2015001 but ranked last on SEED. Since the two variants had identical parameter counts and computational costs, this result indicates that the advantage of Shared SSMoE cannot be attributed solely to increased expert capacity. Instead, sharing the routing space across the spatial and temporal branches appears to promote more robust cross-branch coordination.

The lower-capacity fusion alternatives exhibited different dataset-dependent behaviors. MLP fusion performed relatively well on SEED but was less competitive on BNCI2015001, whereas late fusion maintained moderate performance across both datasets. Bidirectional cross-attention did not consistently outperform the simpler fusion mechanisms, suggesting that introducing more complex token-level interactions does not necessarily lead to better cross-branch representation learning. Overall, these results support the effectiveness of shared expert routing over both independently routed experts and generic cross-branch fusion.

\subsection{SSMoE Expert-Count Sensitivity}
\label{app:ssmoe_capacity}

\begin{figure*}[t]
\centering
\includegraphics[width=0.98\textwidth]{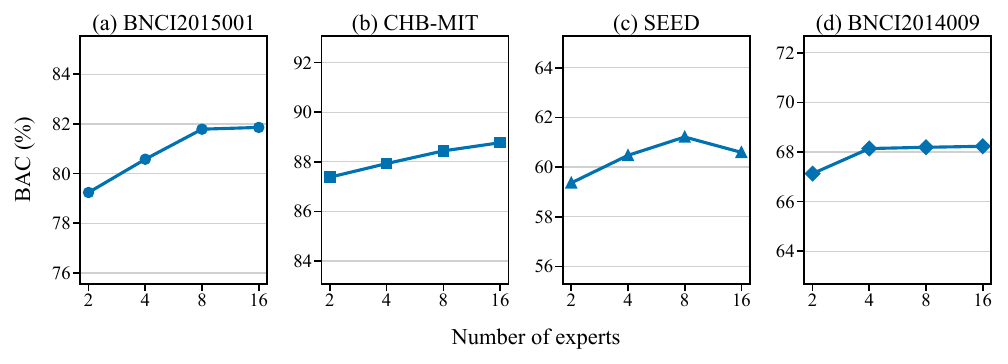}
\caption{Sensitivity of downstream BAC to the number of SSMoE experts under the dataset-specific few-shot protocols.}
\label{fig:expert_sensitivity}
\end{figure*}

To examine the sensitivity of STEAM to the capacity of SSMoE, we varied the number of experts over $\{2,4,8,16\}$ while keeping the remaining architecture, pre-training configuration, and downstream adaptation protocol unchanged. The expert count determines the capacity of SSMoE to aggregate, transform, and redistribute information between the spatial and temporal branches. Figure~\ref{fig:expert_sensitivity} reports the results on four representative downstream datasets under their corresponding few-shot protocols.

Increasing the expert count generally improves downstream performance when the expert pool is small, indicating that insufficient expert capacity can constrain the integration of complementary spatial and temporal representations. As the number of experts further increases, however, the improvements gradually diminish and become more task-dependent. This trend suggests that expanding the expert pool beyond a moderate size introduces additional capacity without consistently yielding proportional performance gains. Overall, the eight-expert configuration achieves strong and stable results across the evaluated datasets, providing an effective balance between representation capacity and model complexity. We therefore adopted eight experts as the default SSMoE configuration.

\subsection{Effect of Training Data Ratio}
\label{app:training_ratio}

\begin{figure*}[t]
\centering
\includegraphics[width=0.98\textwidth]{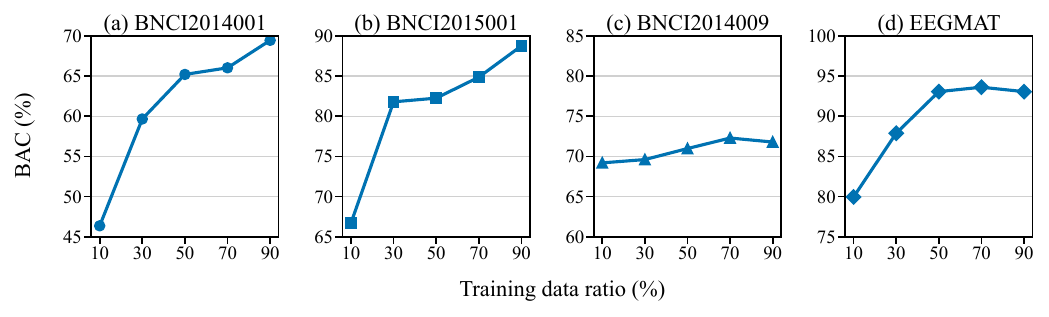}
\caption{Effect of the training data ratio on downstream BAC under full-model fine-tuning.}
\label{fig:training_ratio_curves}
\end{figure*}

We investigated the effect of labeled-data availability by fine-tuning the complete pre-trained model using $10\%$, $30\%$, $50\%$, $70\%$, and $90\%$ of the available training data. All experiments followed the same data-partitioning procedure and full-model fine-tuning configuration, with only the training-data ratio varied. Figure~\ref{fig:training_ratio_curves} presents the resulting BAC curves on four downstream classification datasets.

Downstream performance generally improved as the training-data ratio increased, indicating that STEAM effectively utilized additional task-specific supervision. The largest improvements typically occurred when the ratio increased from small to moderate values, whereas the gains diminished as more labeled data became available. At higher ratios, several curves approached saturation, with minor task-dependent fluctuations rather than strictly monotonic improvements. These results showed that the benefit of additional labeled data depended on both the target task and the sampled training subset. Nevertheless, the overall trend demonstrated that the pre-trained STEAM representations supported effective adaptation across a broad range of training-data ratios and continued to benefit from increased supervision.

\subsection{Subject-Level Results}

\paragraph{Experimental protocol.} Unless otherwise stated, the additional experiments on cross-branch interaction, model capacity, data efficiency, and downstream adaptation followed the same labeled-data protocols as the corresponding Stage-I STEAM experiments reported in the main paper.To ensure a fair and reproducible comparison, all foundation-model baselines are evaluated using their official implementations and publicly released pre-trained checkpoints.

\begin{figure*}[p]
\centering
\includegraphics[width=0.98\textwidth]{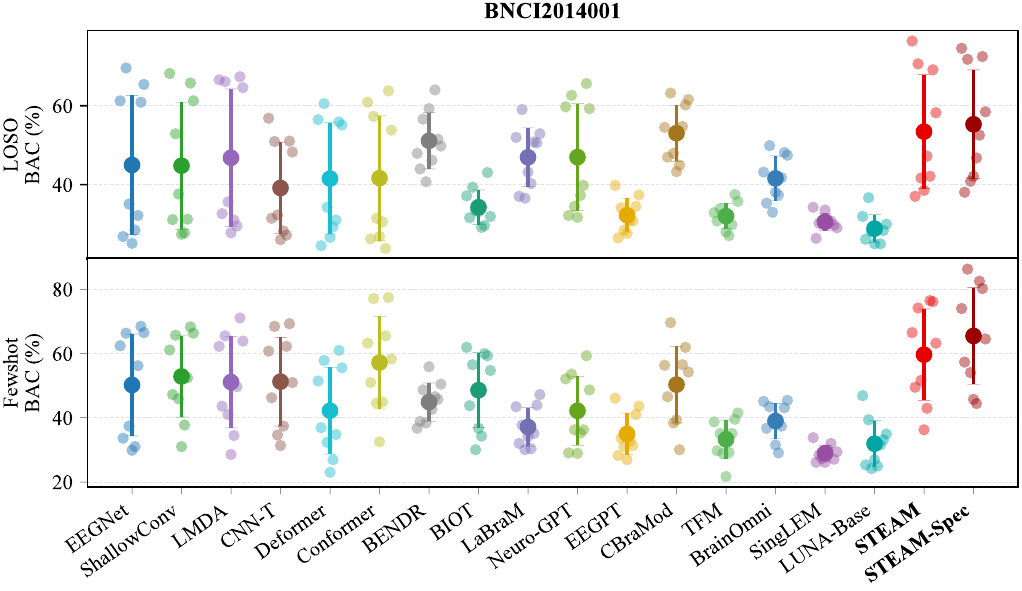}
\caption{Subject-level BAC on BNCI2014001 ($n=9$). The upper and lower panels show LOSO and few-shot performance, respectively.}
\label{fig:subject_bnci2014001}
\end{figure*}

\begin{figure*}[p]
\centering
\includegraphics[width=0.98\textwidth]{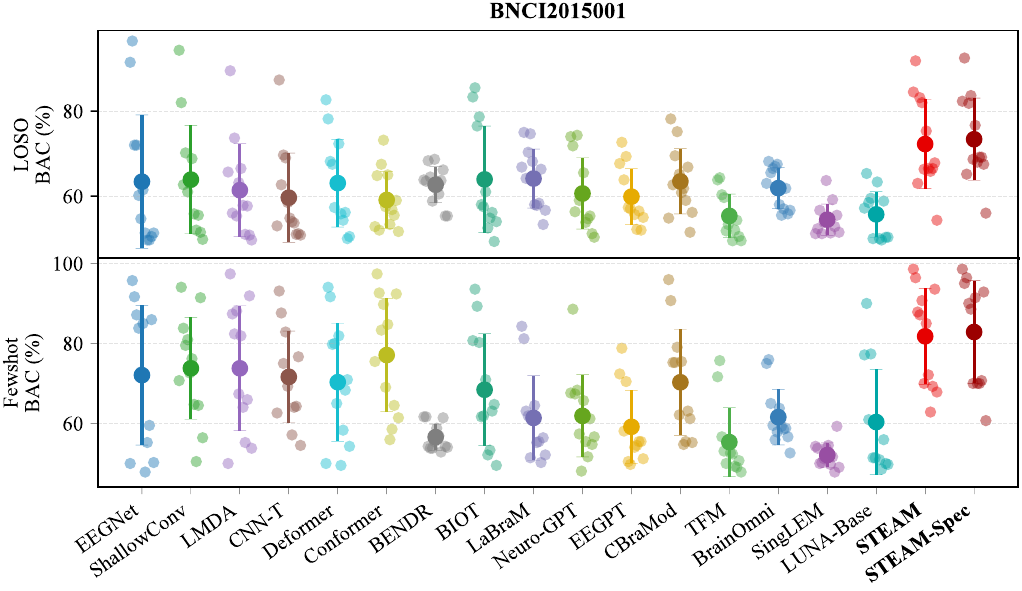}
\caption{Subject-level BAC on BNCI2015001 ($n=12$). The upper and lower panels show LOSO and few-shot performance, respectively.}
\label{fig:subject_bnci2015001}
\end{figure*}

\begin{figure*}[p]
\centering
\includegraphics[width=0.98\textwidth]{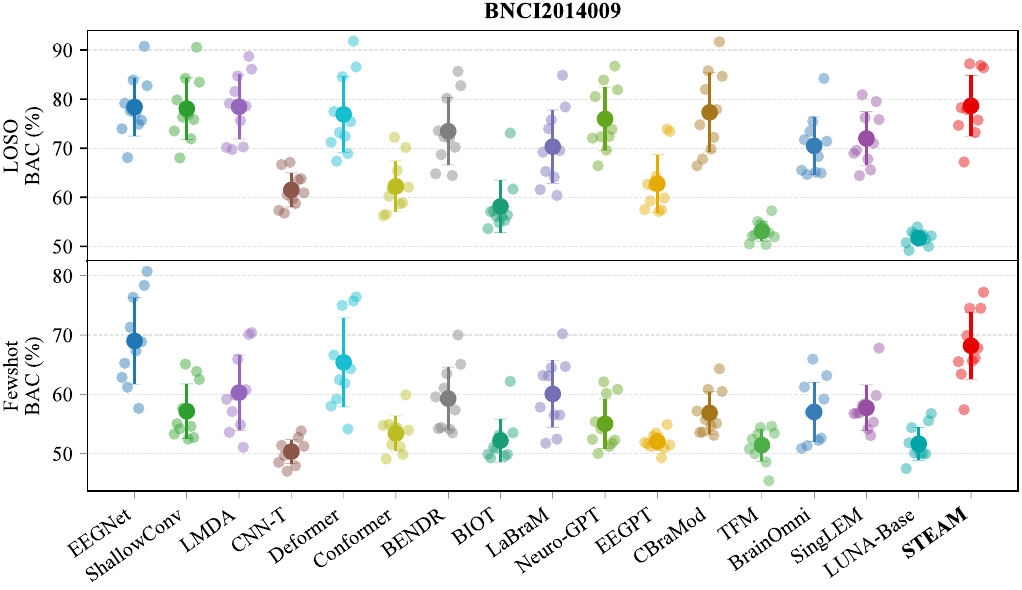}
\caption{Subject-level BAC on BNCI2014009 ($n=10$). The upper and lower panels show LOSO and few-shot performance, respectively.}
\label{fig:subject_bnci2014009}
\end{figure*}

\begin{figure*}[p]
\centering
\includegraphics[width=0.95\textwidth]{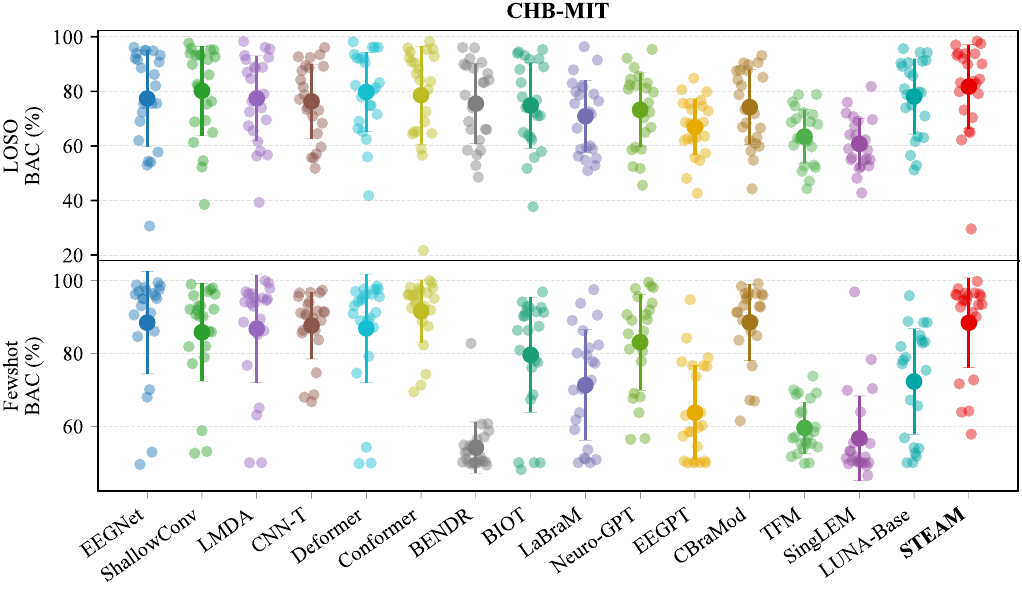}
\caption{Subject-level BAC on CHB-MIT ($n=23$). The upper and lower panels show LOSO and few-shot performance, respectively.}
\label{fig:subject_chbmit}
\end{figure*}

\begin{figure*}[p]
\centering
\includegraphics[width=0.98\textwidth]{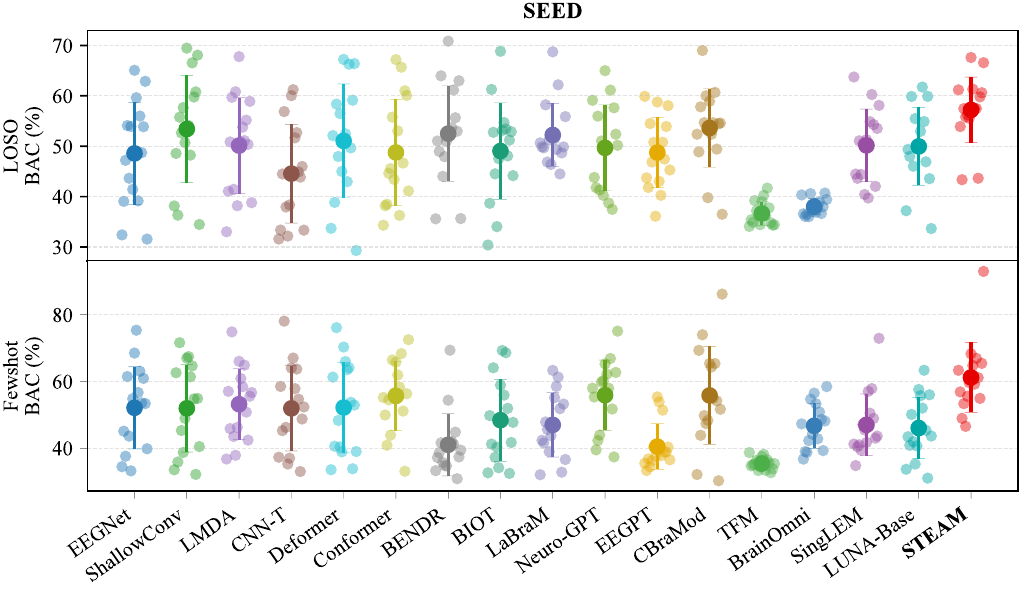}
\caption{Subject-level BAC on SEED ($n=15$). The upper and lower panels show LOSO and few-shot performance, respectively.}
\label{fig:subject_seed}
\end{figure*}

\begin{figure*}[p]
\centering
\includegraphics[width=0.98\textwidth]{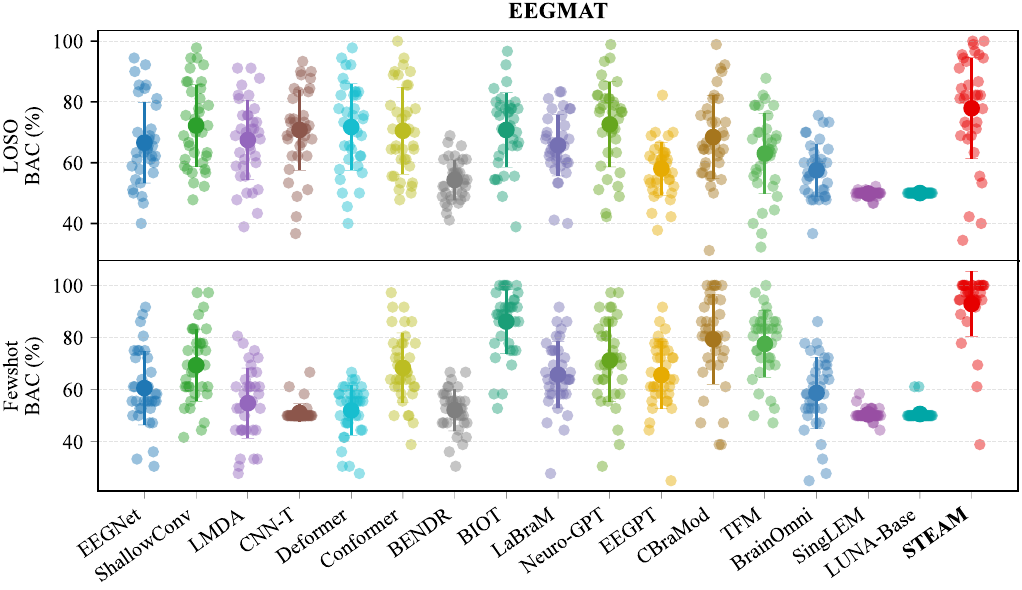}
\caption{Subject-level BAC on EEGMAT ($n=36$). The upper and lower panels show LOSO and few-shot performance, respectively.}
\label{fig:subject_eegmat}
\end{figure*}

\begin{figure*}[!t]
\centering
\includegraphics[width=0.88\textwidth]{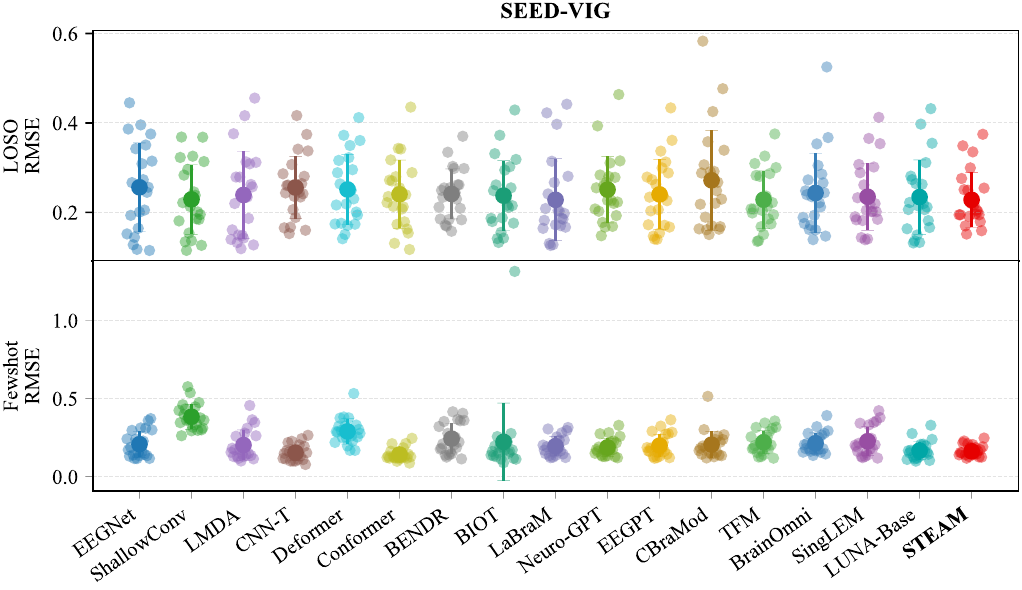}
\caption{Subject-level RMSE on SEED-VIG ($n=21$). The upper and lower panels show LOSO and few-shot performance, respectively; lower values indicate better performance.}
\label{fig:subject_seedvig}
\end{figure*}

Figures~\ref{fig:subject_bnci2014001}--\ref{fig:subject_seedvig} present subject-level performance under the LOSO and few-shot protocols. Translucent markers represent individual-subject scores, while opaque markers indicate the corresponding aggregate results reported in the main paper. Vertical error bars denote the standard deviation across subjects and characterize inter-subject variability. Classification tasks are evaluated using BAC, whereas SEED-VIG is evaluated using RMSE, for which lower values indicate better performance. STEAM denotes the Stage-I general model, while STEAM-Spec denotes the corresponding Stage-II paradigm-specialized variant when specialization results are available.

\paragraph{Aggregate-rank calculation.}
The aggregate ranking includes the six specialist baselines, ten general foundation-model baselines, and Stage-I STEAM reported in the main paper. MIRepNet and STEAM-Spec are excluded because neither reports results for all fourteen dataset--scenario settings. Within each setting, methods are ranked in descending order of BAC or ascending order of RMSE using standard competition ranking. Methods with identical performance are assigned the same rank, defined as one plus the number of methods with strictly better performance, and the subsequent ranks are skipped accordingly. For method $m$, the combined rank is defined as
\[
R_m=|\mathcal{S}_m|^{-1}\sum_{s\in\mathcal{S}_m}r_{m,s},
\]
where $\mathcal{S}_m$ denotes the set of available settings across both the LOSO and few-shot evaluations, and $r_{m,s}$ denotes the rank of method $m$ in setting $s$. BrainOmni is averaged over twelve settings because its CHB-MIT results are unavailable, whereas all other methods are averaged over fourteen settings. Table~\ref{tab:aggregate_ranks} reports the setting-wise and combined ranks of the foundation models included in the computational-efficiency comparison.

Across the LOSO and few-shot scenarios, STEAM achieves a combined rank of 1.43, compared with a mean combined rank of 10.55 for the ten foundation-model baselines, corresponding to a margin of 9.12 rank positions.

\begin{table*}[!t]
\centering
\begingroup
\small
\setlength{\tabcolsep}{1mm}
\renewcommand{\arraystretch}{1}
\begin{tabular*}{\textwidth}{@{\extracolsep{\fill}}lccccccccc@{}}
\toprule
Approach & Scenario & 14001 & 15001 & 14009 & CHB-MIT & SEED & EEGMAT & VIG & Combined \\
\midrule
\multirow{2}{*}{\textbf{STEAM}} & LOSO & 1 & 1 & 1 & 1 & 1 & 1 & 1 & \multirow{2}{*}{\textbf{1.43}} \\
 & Few-shot & 1 & 1 & 2 & 4 & 1 & 1 & 3 & \\
\addlinespace[1pt]
\multirow{2}{*}{CBraMod} & LOSO & 2 & 5 & 5 & 11 & 2 & 8 & 17 & \multirow{2}{*}{6.50} \\
 & Few-shot & 6 & 8 & 10 & 2 & 3 & 3 & 9 & \\
\addlinespace[1pt]
\multirow{2}{*}{Neuro-GPT} & LOSO & 4 & 11 & 7 & 12 & 10 & 2 & 13 & \multirow{2}{*}{8.00} \\
 & Few-shot & 11 & 10 & 11 & 9 & 2 & 5 & 5 & \\
\addlinespace[1pt]
\multirow{2}{*}{LaBraM} & LOSO & 5 & 2 & 11 & 13 & 5 & 11 & 1 & \multirow{2}{*}{8.29} \\
 & Few-shot & 13 & 12 & 5 & 12 & 12 & 8 & 6 & \\
\addlinespace[1pt]
\multirow{2}{*}{BIOT} & LOSO & 13 & 3 & 15 & 10 & 11 & 5 & 7 & \multirow{2}{*}{9.21} \\
 & Few-shot & 8 & 9 & 13 & 10 & 10 & 2 & 13 & \\
\addlinespace[1pt]
\multirow{2}{*}{BENDR} & LOSO & 3 & 8 & 8 & 9 & 4 & 15 & 11 & \multirow{2}{*}{10.50} \\
 & Few-shot & 9 & 15 & 6 & 16 & 15 & 13 & 15 & \\
\addlinespace[1pt]
\multirow{2}{*}{BrainOmni} & LOSO & 10 & 9 & 10 & -- & 16 & 14 & 12 & \multirow{2}{*}{11.50} \\
 & Few-shot & 12 & 11 & 9 & -- & 13 & 11 & 11 & \\
\addlinespace[1pt]
\multirow{2}{*}{LUNA-Base} & LOSO & 17 & 15 & 17 & 5 & 9 & 16 & 5 & \multirow{2}{*}{12.36} \\
 & Few-shot & 16 & 13 & 15 & 11 & 14 & 16 & 4 & \\
\addlinespace[1pt]
\multirow{2}{*}{EEGPT} & LOSO & 14 & 12 & 12 & 14 & 13 & 13 & 9 & \multirow{2}{*}{12.43} \\
 & Few-shot & 14 & 14 & 14 & 13 & 16 & 9 & 7 & \\
\addlinespace[1pt]
\multirow{2}{*}{SingLEM} & LOSO & 16 & 17 & 9 & 16 & 7 & 17 & 6 & \multirow{2}{*}{13.29} \\
 & Few-shot & 17 & 17 & 7 & 15 & 11 & 17 & 14 & \\
\addlinespace[1pt]
\multirow{2}{*}{TFM} & LOSO & 15 & 16 & 16 & 15 & 17 & 12 & 3 & \multirow{2}{*}{13.43} \\
 & Few-shot & 15 & 16 & 16 & 14 & 17 & 4 & 12 & \\
\midrule
Mean of 10 FM baselines & & & & & & & & & 10.55 \\
\bottomrule
\end{tabular*}
\endgroup
\caption{Setting-wise and combined ranks of the general EEG foundation models. Each approach occupies two rows corresponding to LOSO and few-shot evaluation, and the final column averages the ranks over all available settings from both scenarios. Ranks follow standard competition ranking, under which tied methods receive the same rank and subsequent ranks are skipped accordingly. Lower values indicate better overall performance. The abbreviated headers 14001, 15001, and 14009 denote BNCI2014001, BNCI2015001, and BNCI2014009, respectively; VIG denotes SEED-VIG.}
\label{tab:aggregate_ranks}
\end{table*}

\end{document}